\documentclass[review,onecolumn,times]{elsarticle}

\usepackage{booktabs}
\usepackage{natbib}
\usepackage{multirow}
\usepackage{amsfonts}
\usepackage{amssymb}
\usepackage[usenames]{pstricks}
\usepackage{enumitem}
\usepackage{hyperref}
\usepackage{amsmath}
\usepackage{lineno}
\usepackage{stmaryrd}

\newcommand{\x}{\ensuremath{\mathbf{x}}}

\newcommand{\mytitle}{Deep multi-task learning for a geographically-regularized semantic segmentation of aerial images}

\journal{ISPRS Journal of Photogrammetry and Remote Sensing}

\begin{document}
\begin{frontmatter}

\title{\mytitle}
\author[1]{Michele Volpi~\corref{MVgrant}}
\cortext[MVgrant]{Corresponding Author: Michele Volpi, michele.volpi@sdsc.ethz.ch, Tel: +41 (0) 44 635 5256, Fax: +41 (0) 44 635 6848. DT: devis.tuia@wur.nl. Web: https://sites.google.com/site/michelevolpiresearch/}
\address[1]{Swiss Data Science Center, ETH Zurich, Switzerland.\fnref{label3}}
\author[2]{Devis Tuia}
\address[2]{Laboratory of GeoInformation Science and Remote Sensing, Wageningen University and Research, the Netherlands}

\begin{abstract}
\noindent \textbf{This is the pre-acceptance version, to read the final version published in the ISPRS Journal of Photogrammetry and Remote Sensing, please go to: \href{https://doi.org/10.1016/j.isprsjprs.2018.06.007}{10.1016/j.isprsjprs.2018.06.007}}.

When approaching the semantic segmentation of overhead imagery in the decimeter
spatial resolution range, successful strategies usually combine powerful methods
to learn the visual appearance of the semantic classes (e.g. convolutional
neural networks) with strategies for spatial regularization (e.g. graphical
models such as conditional random fields).

In this paper, we propose a method to learn evidence in the form of semantic
class likelihoods, semantic boundaries across classes and shallow-to-deep visual
features, each one modeled by a multi-task convolutional neural network
architecture. We combine this bottom-up information with top-down spatial
regularization encoded by a conditional random field model optimizing the label
space across a hierarchy of segments with constraints related to structural,
spatial and data-dependent pairwise relationships between regions.

Our results show that such strategy provide better regularization than a series
of strong baselines reflecting state-of-the-art technologies. The proposed
strategy offers a flexible and principled framework to include several sources
of visual and structural information, while allowing for different degrees of
spatial regularization accounting for priors about the expected output
structures.
\end{abstract}


\end{frontmatter}


\section{Introduction}

This paper deals with parsing decimeter resolution abovehead images into
semantic classes, relating to land cover and/or land use types. We will refer to
this process as \emph{semantic segmentation}. For a successful segmentation, one
requires visual models able to disambiguate local appearance by understanding
the spatial organization of semantic classes \citep{gould2008ijcv}. To this end,
machine learning models need to exploit different levels of spatial continuity
in the image space \citep{campbell1997pattrec,shotton2006eccv}. Accurate land
cover and land use mapping is an active research field, growing in parallel to
developments in sensors and acquisition systems and to data processing
algorithms. Applications ranging from environmental monitoring
\citep{asner2005science,gimenez2017rse} to urban studies
\citep{zhong2007tgrs,jat2008jag} benefit from advances in processing and
interpretation of abovehead data.

Semantic segmentation of sub-decimeter aerial imagery is often tackled by Markov
and conditional random fields (MRF, CRF)  \citep{besag1974jrss,lafferty2001icml}
combining local visual cues (the \emph{unary} potentials) and interaction
between nearby spatial units (the \emph{pairwise} potentials)
\citep{kluckner2009accv,hoberg2015tgrs,zhong2007tgrs,shotton2006eccv,
volpi2015bcvprw}. By maximizing the posterior joint probability of a CRF over
the labeling (i.e. minimizing a Gibbs \emph{energy}), one retrieves the most
probable \emph{labeling} of a given scene, i.e. the most probable configuration
of local label assignments over the whole image space. These frameworks allow to
model jointly bottom-up evidence, encoded in the unary potentials, together with
some domain specific prior information encoded in the spatial interaction
pairwise terms.

The idea behind the proposed model is that, when dealing with urban imagery (and
in general decimeter resolution imagery), both the content of the image and the
classes are highly structured in the spatial domain, calling for data- and
domain-specific regularization. To follow such intuition, we model two key
aspects of spatial dependencies: \emph{input} and \emph{output} space
interactions. The former are usually encoded by operators accounting for the
spatial autocorrelation of pixels in their spatial domain. The latter are
encoded by different kinds of pairwise potential, favoring specific
configurations issued from a predefined prior distribution.

\begin{itemize}

\item[-] To extract information about local \emph{input} relations, we combine
state-of-the-art convolutional neural networks (CNN, \citep{lecun1998pieee,
simonyan2015iclr, krizhevsky2012nips}) providing data-driven cues for multiple
tasks: We employ a CNN to not only provide approximate class-likelihoods, but
also to predict semantic boundaries between the different classes. The latter
coincide usually with natural edges in the image, but also corresponding to
changes in labeling. Then, we build a segmentation tree using the semantic
boundaries predicted by the CNN. Such tree represents hierarchy of regions
spanning from the lowest level defined by groups of pixels (or superpixels) to
the highest level, the whole scene. The region partitioning depends jointly on
shallow-to-deep visual features \emph{and} the semantic boundaries learned by
the multi-task CNN. 

\item[-] To account for the \emph{output} relations between regions, we combine
the information within each region in a hierarchy using a top-down graphical
model model including different key aspects of the spatial organization of
labels, given the observed inputs. This second modeling step is based on a CRF
that aims at reducing the complexity (i.e. regularizing) of the pixel-wise maps,
by semantically and spatially parsing consistent regions of the image, likely to
belong to given classes, at different scales. Specifically, the CRF model takes
into account evidence from the CNN (class-likelihoods, learned visual features
and presence of class-specific boundaries) and spatial interactions (label
smoothness, label co-occurrence, region distances, elevation gradient) within
the hierarchy. In other words, it learns the extent \emph{and} the labeling of
each segment simultaneously, by minimizing a specifically designed energy.
\end{itemize}

\begin{figure*}[!t] \centering
\includegraphics[width=0.8\textwidth]{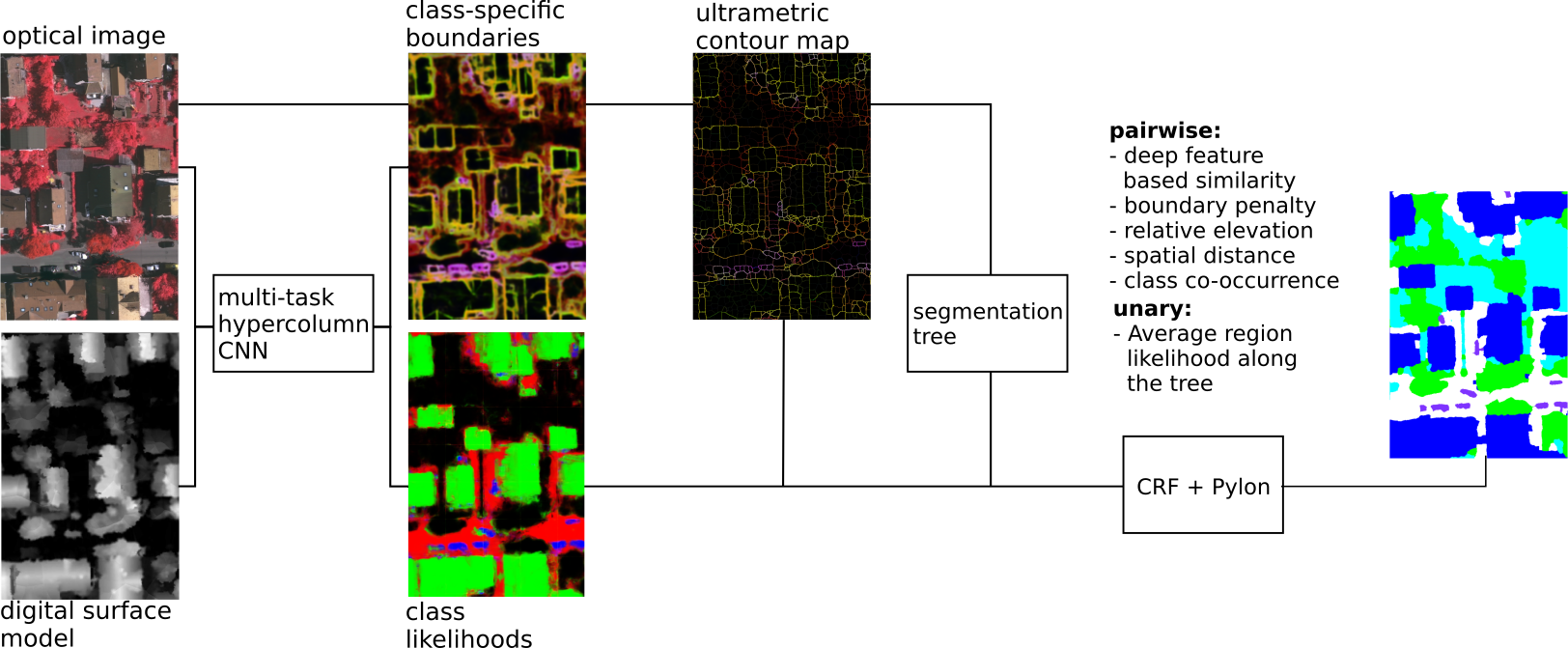}
\caption{Flowchart of the proposed pipeline: from multi-modal inputs and
multi-task learning to multi-modal and geographically regularized semantic
segmentation.}
\label{fig:flowchartpipeline} \end{figure*}

A visual summary of the proposed pipeline is presented in
Figure~\ref{fig:flowchartpipeline}.

We evaluate all the components of the system and show that spatial
regularization is indeed useful in simplifying class structures spatially, while
achieving accurate results. Since spatial structures are learned and encoded
directly in the output map, we believe our pipeline is a step towards systems
yet based on machine learning, but not requiring extensive manual
post-processing (e.g. local class filtering, spatial corrections, map
generalization, fusion and vectorization
\cite{crommelinck2016remsens,hoehle2017remsens}), at the same time employing
domain knowledge and data specific regularization, tailoring it to specific
application domain and softening black-box effects. Specifically, the
contributions of this paper are:
\begin{itemize}
\item[-] A detailed explanation on our multi-task CNN, building on
top of a pretrained network (VGG);
\item[-] A strategy to transform semantic boundaries probabilities to
superpixels and hierarchical regions;
\item[-] A CRF encoding the desired space-scale relationships between segments;
\item[-] The combination of different energy terms accounting for multiple
input-output relationships, combining bottom-up (outputs and features of the
CNN) and top-down (multi-modal clues about spatial arrangement) into local and
pairwise relationships.
\end{itemize}

In the next Section, we summarize some relevant related works. In
Section~\ref{sec:met}, we present the proposed system: the multi-task CNN
architecture (Sec.~\ref{sec:theory-cnn}), the hierarchical representation of
image regions (Sec.~\ref{sec:theory-ucm}) and the CRF model
(Sec.~\ref{sec:crf}). We present data and experimental setup in
Section~\ref{sec:data-setup} and the results obtained in
Section~\ref{sec:results}. We finally provide a discussion about our system in
Section~\ref{sec:discussion}, leading to conclusions presented in
Section~\ref{sec:concs}.

\section{Related works}

\subsection{Mid-level representations}

To generate powerful visual models, traditional methods compute local appearance
models mapping locals descriptors to labels, over a dense grid covering the
image space. Then, the relationships between output variables are usually
modeled by MRF and CRF. Standard approaches to local image descriptors involve
the use of local color statistics, texture, bag-of-visual-words, local binary
patterns, histogram of gradients and so on
\citep{kluckner2009accv,hoberg2015tgrs,zhong2007tgrs,volpi2015bcvprw,shotton2006eccv}.

However, the use of fixed shape operators causes an inevitable loss of
geometrical accuracy, in particular on objects borders. The most common solution
to this problem is to employ a locally adaptive spatial support, to either
summarize precomputed dense descriptors or to retrieve new ones. This strategy
is usually implemented by the use of superpixels, which are defined to be small
spatial units, uniform in appearance, while matching the natural edges in the
image \citep{felzenszwalb2004ijcv}. Moreover, superpixels significantly reduce
the number of atomic units to be processed and consequently the computational
time.

\subsection{Deep representations}

Convolutional neural networks (CNN), learn a parametric mapping from inputs to
outputs, sidestepping the definition of i) an explicit processing resolution,
ii) the type of appearance descriptors best representing the data, iii) a
classifier to map descriptors to output labels \citep{lecun1998pieee,
simonyan2015iclr, krizhevsky2012nips}. The fact that a CNN learns an end-to-end
mapping from data to outputs directly \emph{within} the network makes deep
neural networks more than valid alternatives to classical models, since complex
feature engineering is avoided. However, most of these methods still rely on a
fixed spatial support, either in the form of a patch to be classified (for patch
classification) or in the form of the convolution kernel (for fully
convolutional architectures) \citep{long2015cvpr}. This can cause boundary blur
and loss of definition on small image details. But despite blurring effects
related to the size of the field of view, fully convolutional architectures are
of particular interest for semantic segmentation. 
These techniques provide impressive accuracy and, once the architecture and
hyperparameters are set, can be trained with ease as long as enough training
data is available to learn the parameters.

For aerial and satellite image semantic segmentation, fully convolutional
architectures are becoming state-of-the-art. A notable work in this direction is
by \citet{sherrah2016arxiv}, where the author modifies a pretrained network  to
not downsample and performs dense semantic segmentation directly. In our
previous work \citep{volpi2017tgrs}, we explored the use of deconvolutions to
learn upsamplings providing outputs of the same resolution as inputs. Results
outperformed standard patch-based strategies, both in terms of accuracy metrics
and computational time, suggesting that fully convolutional architectures are
best suited for semantic segmentation. In \citep{marcos2018jisprs} we extended
standard CNN to incorporate rotation invariance at the convolutional kernel
level, resulting in small and compact models, but providing accuracy comparable
to state-of-the-art architectures. \citet{maggiori2017tgrs} showed that a
slightly modified \emph{hypercolumn} architecture (originally proposed in
\citep{hariharan2015cvpr}) is well suited for semantic segmentation of aerial
images. Hypercolumns architectures exploit the entirety of the multiscale
structures learned within the network, by upsampling intermediate activations
and stacking them into a \emph{hypercolumn layer}. On top of the latter, another
sub-network learns to map to desired outputs, by exploiting the stack of
multiscale structures, that can be seen as input features to this sub-network,
and learning how to combine them. This is desirable for a model working in a
highly structured input-output space with unique and explicit spatial relations.
Due to these properties, we also adopt a modification of the hypercolumn
architecture, which we detail in Section~\ref{sec:theory-cnn}.

\subsection{Learning boundary detectors}

In this work we propose a multi-task CNN architecture to jointly learn a
semantic segmentation task and a semantic boundary detection task. This latter
task relates to supervised learning of semantic boundaries, as introduced
in \citep{hariharan2011iccv}. Authors proposed to combine bottom up edge
detectors with object detectors in the image to enhance boundaries semantically
The framework presented in the latter work has been reformulated in terms of
deep learning \citep{shen2015cvpr,xie2015cvpr,kokkinos2016iclr,
maninis2017tpami}.

Usually, the prediction of boundaries supports semantic segmentation tasks, as
the edges are naturally embedding shape information and contours of semantic
classes. Such ideas have been exploited in computer vision since years (see e.g.
the works stemming from \citep{arbelaez2011tpami}) and now also framed as deep
learning systems \citep{kokkinos2017cvpr,maninis2017tpami} or
\citep{marmanis2018jisprs} for an example in aerial image segmentation.

In this work we build on the idea to retrieve semantic boundaries for one
specific semantic class at a time. In this respect, we implement a sort of
situational boundary detector \citep{uijlings2015cvpr}, but rather than training
a boundary detector on groups of images with similar appearance we train it on
binary subproblems. More specifically, we train our detector to learn
boundaries of one class at a time, versus non-edges for that class. For
instance, we would learn a building versus non-building boundary detector,
resulting in only building contours to be learned.

\subsection{Models of output structures}

Most CRF formulations model the conditional smoothness of the labels in the
output space \citep{boykov2001tpami}, facilitating two neighboring locations to
share the same label if the mutual local appearance is similar. It is also
common to incorporate information about the prior distribution of the label by
weighting this potential with information about the statistical co-occurrence or
relative location of labels at neighboring locations \citep{gould2008ijcv}.

In \citet{volpi2015bcvprw} the energy employed to perform semantic segmentation
is composed by a contrast-sensitive spatial quantization of label interactions
allowing to learn specific dependencies at different scales with a structured
Support Vector Machine. \citet{hoberg2015tgrs} present an approach not only
modeling spatial dependencies at different scales, but also including
class-specific temporal transition probabilities. \citet{hedhli2016tgrs} propose
a hierarchical CRF to model image time series of different spatial resolution,
where the hierarchy copes with the different scales represented in the image.
\citet{zheng2017tgrs} propose a semantic segmentation system exploiting a
hierarchy of labels in order to flexibly deal with possibly different labelings
of the image. The auxiliary labels for each region in the images are modeled by
a CRF. \citet{golipour2016tgrs} exploit a hierarchy of regions built by
region-merging into a MRF model. They also extrapolate a score over edges of
regions by checking relationships in the hierarchy. Finally, \citet{li2015tgrs}
rely on a higher-order CRF to jointly infer at pixel and region level whether
pixels and regions belong to rooftops or not. The line followed by these works
show that accounting for hierarchical relationships jointly with the spatial
distribution of labels allows for more expressive models.

In this paper we model spatial dependencies by following this last line of
works. We employ a segmentation tree to encode multilevel likelihoods, in which
regions merged according to learned and input cues should approximate object
extent, and thus favoring a geometrically and semantically accurate
segmentation.

\section{Deep parsing of aerial images}\label{sec:met}

Our model is composed of three main ingredients: a multi-task CNN providing
class-likelihoods and probabilities of boundaries
(Section~\ref{sec:theory-cnn}), a segmentation tree
(Section~\ref{sec:theory-ucm}) and a CRF model encoding information about
spatial dependency of the labeling (Sect.~\ref{sec:crf}).

\subsection{Multi-task CNN}\label{sec:theory-cnn}

\begin{figure}[!t]
\centering
\begin{tabular}{ccc}
\multicolumn{2}{c}{\includegraphics[height=0.23\textheight]{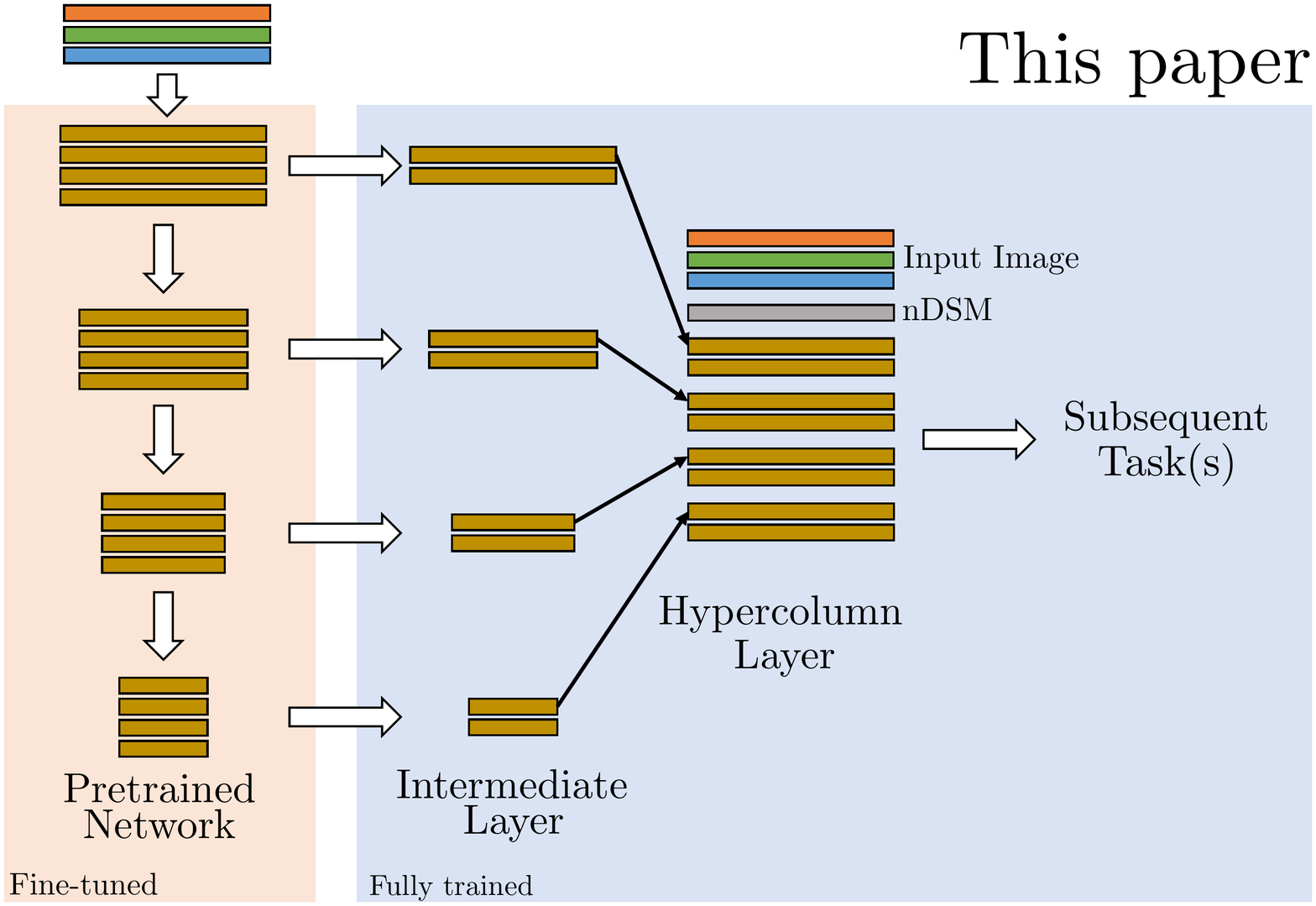}}\\
\multicolumn{2}{c}{(a)}\\
\includegraphics[height=0.23\textheight]{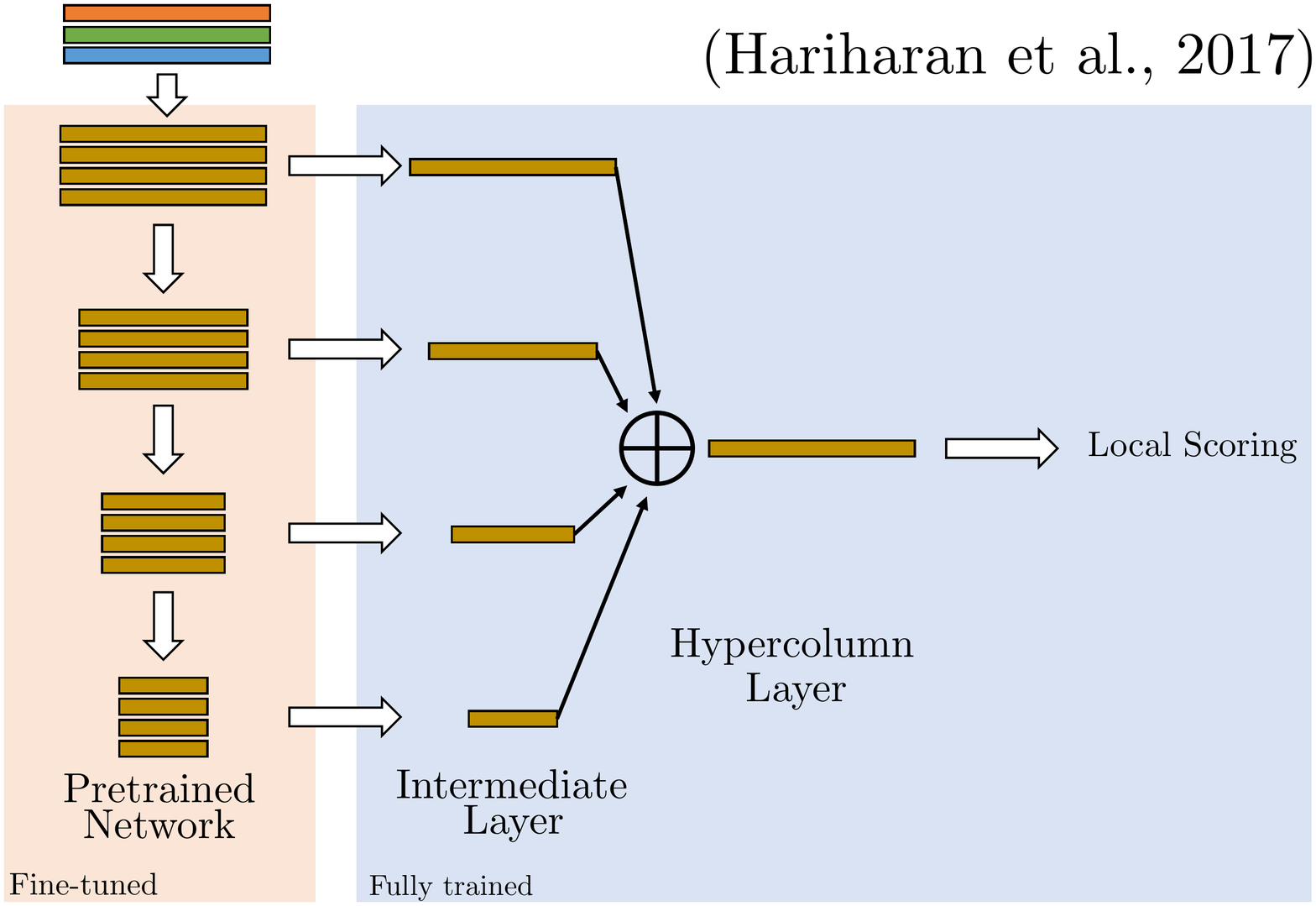} &
\includegraphics[height=0.23\textheight]{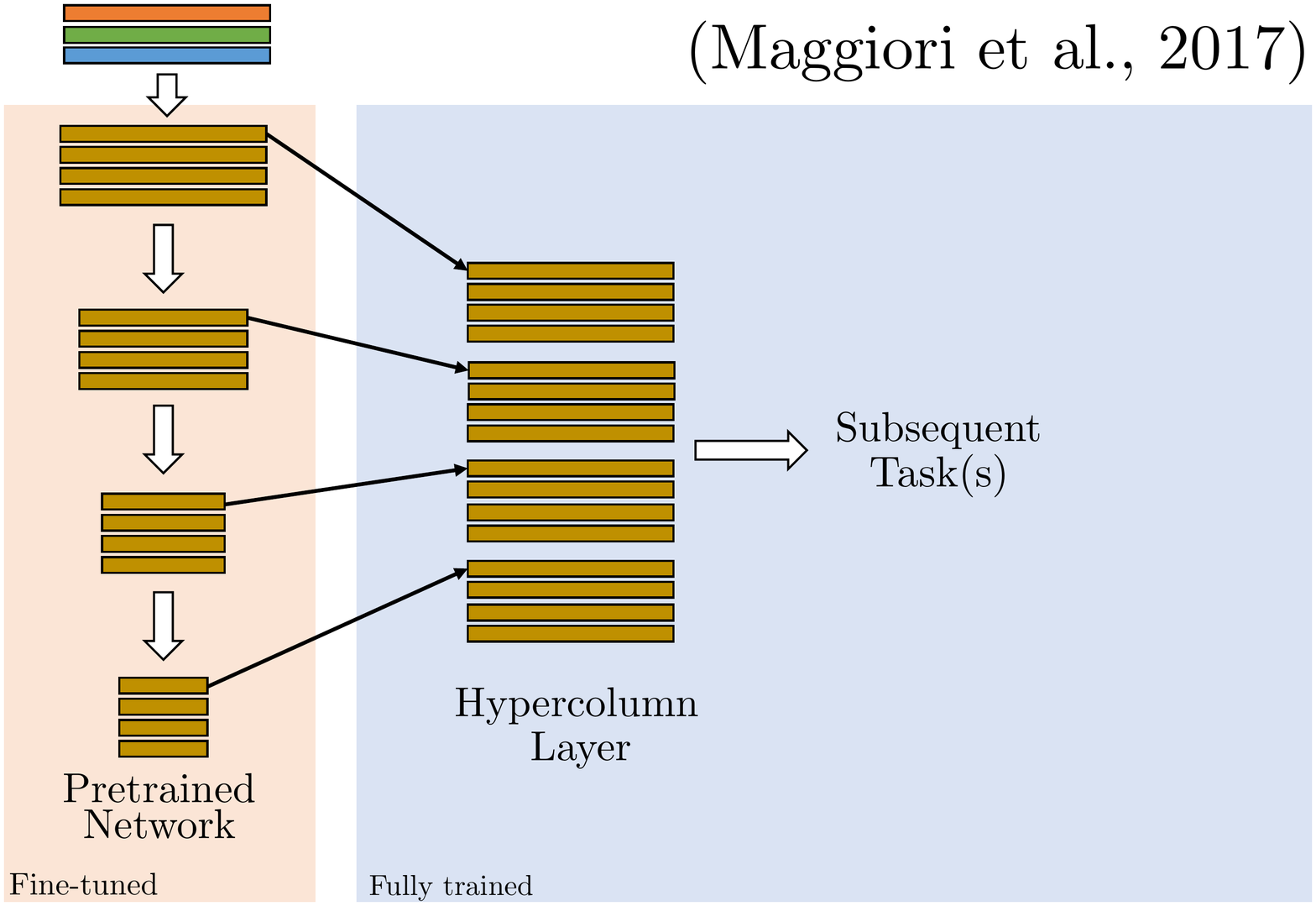} \\
(b) & (c) \\
\end{tabular}
\caption{A comparison of three hypercolumn architectures. In all images, white empty arrow means learned convolutions while thin arrows correspond to bilinear upsampling. (a) shows our architecture, (b) the one from \citep{maggiori2017tgrs} and (c) the original hypercolumn as presented by \citep{hariharan2015cvpr}. Note that this last one, since sums responses from activation-specific classifiers, cannot be used for multitask learning out of the box.}
  \label{fig:architecture}
\end{figure}

Figure~\ref{fig:architecture}(a) presents the general network architecture. It
closely follows the hypercolumn architectures presented in
\citep{hariharan2015cvpr} (illustrated in Figure~\ref{fig:architecture}(b)),
then used in \citep{maggiori2017tgrs} (Figure~\ref{fig:architecture}(c)), yet
with small differences from both.

The main trunk of the network consists in a standard classification network,
specifically the VGG-16 network pretrained on the ImageNet Image Large Scale
Visual Recognition Challenge (ILSVRC) \citep{simonyan2015iclr}, which maps fixed
size inputs to class scores. However, differently from other works employing
pretrained networks, we remove the last fully connected layers and the last
convolutional block and add two additional task-specific sub-networks to
learn semantic segmentation and semantic boundary detection jointly. We do so,
after rearranging the whole structure to follow the hypercolumns strategy
\citep{hariharan2015cvpr}.

Deep neural networks learn mappings with increasing abstraction power: from
inputs to deep layers the learned features span from general and basic image
descriptors (e.g. gradient detectors) to semantic concepts. The hypercolumn
strategy aims at making use of such hierarchical abstraction in a more general
framework, supporting fully convolutional architectures, and therefore no longer
dependent on the extent of spatial activations.

To this end, such architectures resample activations from each block and we
learn a mapping to the \emph{hypercolumn} layer. On top of this hypercolumn, one
or more convolutional layers learn a mapping into class scores. The rationale
behind is that, by combining different levels of information, the first fully
connected layer is not exclusively dependent from the activations learned
sequentially up to that point, but can reuse previous activations if the problem
requires so. This idea has been further extended in the DenseNet architectures
\citep{huang2017cvpr}.

Differently from standard hypercolumn models, we jointly learn two tasks on top
of the hypercolumn stack: a semantic segmentation and a semantic boundary task.
To this end, the network learns jointly two independent branches composed by two
layers and a class score layer each (one per task). The semantic segmentation
and the semantic boundary tasks are learned by minimizing a standard weighted
cross-entropy loss, but with different scores aggregation strategies. This
choice follows logically from the fact that a given pixel can belong to several
semantic boundaries at the same time, typically two or three, while usually can
only belong to a single semantic class.

The global loss function is given by a linear combination of the independent
losses for each task $t$ as:
\begin{equation} \mathcal{L}(y,\hat{y}) = \sum_{i\in\{S,B\}} \beta_t \mathcal{L}^i(y,\hat{y}),
\end{equation}
where $\beta$ balances the contribution of each loss. We observed that changing
this hyperparameter modifies the speed of convergence of a task at the expense
of the other. Loss scaling does not seem to be a problem as each multi-task
branch is independently learned. Therefore, we set the contribution of both
losses to be equal.

Both functions are defined by cross-entropy losses: the semantic segmentation
loss $\mathcal{L}^S$ as:
\begin{equation} \mathcal{L}^S(y,\hat{y}) = - \frac{1}{N} \sum_c \omega_c \llbracket y_i = c \rrbracket \
\log \big(p(y_i = c | x_i)\big),
\label{eq:crossentropy}
\end{equation}
where $N$ is the number of pixels in every input patch, $y_i$ is the ground
truth class assigned to pixel $i$ and $\hat{y}_i$ the prediction. The posterior
$p(y_i= c|x_i)$ is obtained by normalizing the activations through a sigmoid.
The cross-entropy loss puts all the mass on a single class $c$, meaning that an
increase in the posterior probability for one class corresponds to a decrease in
another one.

Regarding learning the semantic boundaries, we define the loss $\mathcal{L}^B$
as the average cross-entropy loss computed for each semantic boundary
subproblem, as:
\begin{align}
\mathcal{L}^B(y_i,\hat{y}_i) & = \sum_b
\mathcal{L}^b(y^b_i,\hat{y}^b_i) \nonumber\\ & = - \frac{1}{N}  \sum_b \omega_b
\llbracket y^b_i = b \rrbracket \ \log \big(p(y^b_i = b | x_i)\big),\label{eq:Lb}
\end{align}
where $y^b_i$ and $\hat{y}^b_i$ are the binary ground truth and prediction for
the $b$th binary semantic boundary problem for pixel $i$. The posteriors
$p(y^b_i = c |  x_i)$ are again obtained by normalizing the activations
through a sigmoid.

Therefore, the final loss allows a given pixel to be predicted as boundary
across several classes. Recall that, since we approach modeling semantic
boundaries as separate sub-problems, every $b$th task involves modeling only the
semantic boundaries of class $b$, pooling non-boundary pixels and boundaries
from other classes.

Both losses are weighed for each class by $\omega$: for the segmentation loss,
we set a weight proportional to the inverse class frequency
\citep{mostajabi2015cvpr}, while for boundary we weigh errors on boundaries by
0.99, while on non-boundaries 0.01, since the prior probability of boundary
pixels is extremely low \citep{xie2015cvpr}.

During backpropagation, only the partial derivative of the total loss of
Eq.~\eqref{eq:crossentropy} with respect to the weight for one task is non-zero.
Therefore, only the sub-loss computed at one output backpropagates gradients for
a specific task. Gradients at the hypercolumn layer are added and backpropagated
through the intermediate layers and added again to those coming from the layers
of the main trunk. Learning rates are scaled by a factor $10^{-3}$ on the layers
pretrained on the ILSVRC ImageNet.

Differently from \citep{hariharan2015cvpr}, we do not sum activations at the
hypercolumn layer, since in their model each intermediate layer is a classifier.
Although this is a very effective way for single tasks, in our model we needed
more flexibility to learn both tasks jointly. To this end, we stacked the input
image to activations derived from the rectified linear units nonlinearities
(ReLU) of the classification network. This composes the input to both fully
connected layers, composed by two layers. On the other hand, our model does not
simply stack resampled activations, but also learns an intermediate mapping from
the ReLU to the hypercolumns by a convolutional block (see
Figure~\ref{fig:architecture}). We observe this strategy to be beneficial for
three main reasons: first, it allows the hypercolumn layer to be composed by
slightly altered activations, closer to the tasks that we want to learn. This is
especially important when using pretrained networks, since, although being
related, tasks and modality of our problem differ to those of the ImageNet
ILSVRC. Secondly, we map arbitrarily sized mid-representations from ReLUs to
activations of same dimensionality, typically smaller or equal to the
dimensionality of the original ReLU activations. Although this might seem
unimportant since we stack activations (contrarily to~\citep{hariharan2015cvpr}
where the hypercolumn is a sum of activations), it still has at least three main
benefits:
\begin{itemize}
 \item[-] It allows to control the dimensionality of the hypercolumn stack and
the total number of learnable parameters
 \item[-] We distill information closely related to the tasks at hand by keeping
 the added layers lower-dimensional
 \item[-] Intermediate features also learn important local correlations in the
 feature maps, since these intermediate mappings are learned by standard
 convolutional layers.
\end{itemize}

When coupled to the effective field of view of the classification network and to
the spatial resampling needed to stack activations into the hypercolumns, the
fully connected layers have at their disposal information about features
carrying different levels of abstraction and accounting for different degrees of
spatial dependencies, both short- and long-range. Note that at this level we
also include the original optical image and information about the elevation (in
the form of a normalized digital surface model), so effectively coupling the
power of large scale pretrained network to extract features and additional
information which could not be fed to it.

\subsection{From probability of semantic boundaries to segmentation trees}\label{sec:theory-ucm}

In the presented pipeline, we use the learned semantic boundaries at two
different levels in our pipeline: as a pairwise potential in the CRF model (see
Eq.~\eqref{eq:phib}) and as features to retrieve the segmentation tree, whose
delineation is explained in this section.

In order to obtain a segmentation tree common to all classes, we combine the
semantic boundary probabilities into a score $g$. This score denotes the maximum
posterior probability for a pixel $\x_i$ to belong to a semantic boundary in
task $\mathcal{L}^B$ (Eq.~\eqref{eq:Lb}): $$g(\x_i) =
\max_{b}\left(p(y^b_i|\x_i)\right).$$ This ensures that, independently of the
semantic boundary class, strong edges always prevail over weak ones, enabling us
to obtain the most probable segmentation tree. We will employ the learned
class-specific boundaries in the CRF model.

Before building the segmentation tree, we transform the boundary scores $g(\x)$
into an \emph{ultrametric contour map} (UCM, \citet{arbelaez2011tpami}). An UCM
defines a hierarchy of non-overlapping and closed contours based on a boundary
probability score. The basic level of the hierarchy is defined by region
contoured by low-probability boundaries, which in our case is defined by a
watershed oversegmentation, where the landscape is defined by $g(\x)$. Moving up
in the segmentation tree, only boundaries with stronger UCM contours define
closed contours. The attractive property of UCMs is that, for a given edge
between two regions, the score is constant and defines a natural dissimilarity
metric across regions.

In order to build the segmentation tree, we complement the UCM dissimilarity
with other two metrics defined over regions: the Euclidean distance between
per-region average hypercolumn activation $\x^h$ and the Euclidean distance
between region centroids $\x^c$ in the spatial domain. The former provides
high-level information about region contents, while the second favors small
isolated regions to be clustered to a nearby neighbor. For any two regions
$i,j$, these distances are defined as $\mathbf{D}_{ij} = \| \x_i - \x_j\|_2$ by
employing the appropriate feature vector in $\x$. If we define these three
dissimilarities as $\mathbf{D}^g$, $\mathbf{D}^h$, $\mathbf{D}^c$ for the UCM,
hypercolumn and spatial distance, respectively, our final segmentation tree is
defined by clustering regions based on a convex combination ($\sum_i w_i = 1$):
$\mathbf{D}^* = w_1\mathbf{D}^g + w_2\mathbf{D}^h + w_3\mathbf{D}^c$. The type
of distances reflect those used as pairwise potentials. Their normalization and
the weights $w$ are selected based on the average class purity per leaf region
on validation images. We observed that the UCM dissimilarity tends to receive
the highest weight, followed by the hypercolumn distances and spatial distance.
Note that we will also use these dissimilarities in the CRF model described
below.

\subsection{Semantic segmentation model}\label{sec:crf}

Our CRF model is defined over the regions of the segmentation tree, and not over
pixels directly. The main assumption behind this choice is that every pixel is
covered by a set of regions with different sizes and somewhere, along the tree,
the region which best explains the semantic land-cover classes is present. Given
this assumption, we naturally chose to employ the Pylon model
\citep{lempitsky2011nips}. The Pylon model is an inference framework that, given
a segmentation tree and relationships across leaf regions, finds the most
probable tree cut and segmentation. This is done by exploiting
pairwise spatial smoothness terms and local evidence about class likelihoods
together. Such joint optimization allows flexibility to choose whether large
segments are better than superpixels to describe a given region or if smaller
ones are to be preferred in order to respect objects boundaries.

We model the joint probability of a labeling given the data as a Gibbs
distribution $p(y|\x) \approx \exp\left(-E(y;\x)\right)$. To obtain the most
probable label assignment (maximum-a-posteriori inference), one has to find
among all possible label permutations for every possible semantic segmentation
map, the one minimizing energy $E(y;\x)$.

We define the energy over the pairwise graph $\mathcal{G} =
(\mathcal{V},\mathcal{E})$, where every $i \in \mathcal{V}$ denotes a node (i.e.
a region) with neighbors $j \in \mathcal{E}_i$. Specifically, we define the CRF
as:
\begin{equation}
E(y;\x) = \sum_{i \in \mathcal{V}} \Big( \varphi(y_i;\x_i) + \sum_{j \in \mathcal{E}_i} \mu(y_i,y_j) \phi(y_i,y_j;\x_i,\x_j) \Big).\label{eq:EN}
\end{equation}
In the above model $\varphi(y_i;\x_i)$ is the cost of including in the labeling
region $i$ with label $y$, given some evidence $\x$, while the term
$\phi(y_i,y_j;\x_i,\x_j)$ is a combination of four different pairwise potentials
and it takes into account the interactions between different labelings of two
neighboring regions $i,j$. The pairwise function $\mu(y_i,y_j)$ provides a term
accounting for global compatibility of labels. The rest of this section details
all these terms.

\paragraph{Unary - $\varphi(y_i,\x_i)$} The unary potential is the negative
log-likelihood provided by the CNN using task $\mathcal{L}^S$v
(Eq.~\eqref{eq:crossentropy}) for each region in the segmentation tree. First,
the dense per-pixel prediction of the CNN is averaged over leaf regions, for
each class. This provides a rough approximation of the class-likelihood
for each region, given their appearance modeled by the CNN. However, since we
are dealing with a segmentation tree, the larger the region the more entropic
the average probability, due to larger noise and variability. Therefore, to
allow to select also parent regions, we weigh each unary by the size (in pixel)
of its region, after transformation to standard scores. We define the unary
potential as:
\begin{equation}\
  varphi(y_i;\x_i) = -\log \left( a_i^\gamma p(y_i|\x_i)\right),
\end{equation}
where $a_i$ denotes the area of region $i$ and $\gamma$ is a parameter
controlling the spread of the values. 

\paragraph{Interaction terms - $\phi(y_i,y_j;\x_i,\x_j)$} The interaction terms
account for a series of prior beliefs about the spatial organization of classes
and their interactions. In this work, this term is a sum of the four terms
below, weighted by $\boldsymbol\lambda$:

\begin{enumerate}
\item \emph{Label smoothness.} This potential encourages neighboring nodes to
share the same label if the visual appearance of the regions is similar. It is
usually based on a similarity across edges, so that connected nodes having
similar colors but different labels penalize the energy more than two regions
having different color, and therefore a low similarity.

In our formulation we do not only use the input color similarity, but also deep
features learned specifically for the problem. To this end, we average the
activations stacked in the hypercolumn layer for each region. Remind that the
hypercolumn layer contains information ranging from the input image to higher
level visual concepts. If we denote $\x^h$ as the set of average features from
the hypercolumn layer, the label smoothness pairwise energy is defined as:
\begin{equation}
 \phi^h(y_i,y_j;\x^h_i,\x^h_j) = \exp\left(\frac{-\| \x^h_i - \x^h \|^2_2}{\sigma^h}\right).\label{eq:phih}
\end{equation}
The hyperparameter $\sigma^h$ is set as the median squared Euclidean distance
between features of all the connected regions.

\item \emph{Edge Penalty.} The edge penalty favors neighbors to share a similar
label if no strong semantic edge separates the regions. This potential relies on
the notion that two given adjacent regions can be very different in color, but
belonging to the same semantic class. The information brought by learned
boundaries aims at compensating such situations, by bringing new bottom-up
evidence about label smoothness, this time accounting for local geometric
features. We define this potential as:
\begin{equation}
  \phi^g(y_i,y_j;\x^g_i,\x^g_j) = \exp\left(\frac{-       \theta(\x^g_i,\x^g_{j})}{\sigma^g}\right),\label{eq:phib}
\end{equation}
where $\theta(\x^g_i,\x^g_{j})$ is the UCM score on the separating edge across
two neighboring regions. As for the label smoothness potential, $\sigma^{g}$ is
set as the median UCM score across neighbors.

\item \emph{Spatial distance.} This potential considers spatial distances
between region centroids. Leaf regions can be of different size. Usually, small
regions whose centroids are close to each other tend to belong to the same
class, while large regions with far apart centroids tend to belong to different
classes. This potential scales the energy with respect to the spatial distance
across regions, forcing smoothing to be stronger for close-by regions and softer
for regions with far apart centroids. Its effect is to avoid having small
isolated regions, independently on color or edge strength. This potential is
defined as:
\begin{equation}
  \phi^c(y_i,y_j;\x^c_i,\x^c_j) = \exp \left(\frac{-\|\x^c_i - \x^c_i\|^2_2}{\sigma^c}\right),\label{eq:phic}
\end{equation}

\item \emph{Elevation difference.} This potential relies on the assumption that
regions belonging to the same object will share similar elevation range:
\begin{equation}
  \phi^e(y_i,y_j;\x^e_i,\x^e_j) = \exp \left(\frac{-|\x^e_i - \x^e_i|_1 }{\sigma^c}\right),\label{eq:phie}
\end{equation}
where $\sigma^e$ is the median of the absolute difference (denoted as the
$|\cdot|$ in the numerator). This potential is actually independent from the
actual absolute elevation $\x^e$, since only relative local differences are
considered. Also, note that the \emph{label smoothness} potential already
contains elevation information, where global region similarity is evaluated by
considering the whole hypercolumn stack. Nevertheless, we opted for making such
information more explicit by adding a dedicated, independently weighted
potential.
\end{enumerate}

Summing up all of the above, the final total pairwise potential
$\phi(y_i,y_j;\x_i,\x_j)$ is given by the weighted sum of each independent
potential, as:
\begin{multline}
  \phi(y_i,y_j;\x_i,\x_j) =  \lambda_h \phi^h(y_i,y_j;\x^h_i,\x^h_j) + \lambda_g \phi^g(y_i,y_j;\x^g_i,\x^g_j) \\
  +  \lambda_c \phi^c(y_i,y_j;\x^c_i,\x^c_j) +  \lambda_e\phi^e(y_i,y_j;\x^e_i,\x^e_j)\label{eq:sumPhi}
\end{multline}

\paragraph{Global label compatibility - $\mu(y_i,y_j)$}

To complete Eq.~\eqref{eq:EN}, and to employ a more informative model that the
standard Potts, we weigh the sum of the pairwise potentials of
Eq.~\eqref{eq:sumPhi} by a co-occurrence score, biasing labels association
depending on the pairwise co-occurrence probability observed in training data.
This statistic carries important information about the spatial organization of
classes and conveys semantic ordering into the model. For instance, it is very
likely to have cars and streets spatially co-occurring, while it is improbable
to have many cars being direct neighbors of water regions.

We estimate this statistic by decomposing the ground truth of the training
images into superpixels (corresponding to the lowest level of the segmentation
tree) and simply counting co-occurrences of labels for adjacent regions. As
superpixels, we use the leaves in the segmentation tree described in
Section~\ref{sec:theory-ucm}.

\begin{figure}
  \centering
  \includegraphics[width=0.3\textwidth]{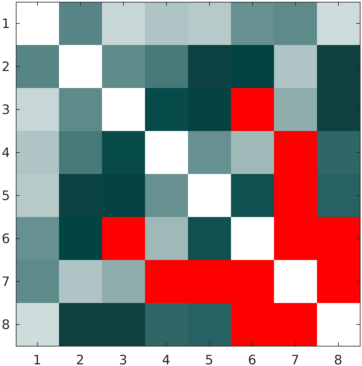}
  \caption{Example of compatibility label matrix between the classes of the
  Zeebrugges dataset (the class IDs are given in Section~\ref{sec:dataze}). Red
  cells denote co-occurrence of classes never observed in training, such as
  class pairs 7 (``Boats'') and 4 (``low vegetation''), 5 (``buildings''), 6
  (``trees'') and 8 (``cars''). On the contrary, lighter colors denote more
  likely combinations, such as 1 (``impervious surfaces'') and 8 (``cars'').}
\label{fig:coocc} \end{figure}

Although estimating such statistic on segmented training images, overfitting is
not an issue, as we only retrieve counts to parametrize a metric with a number
of parameters equal to $C^2 - C$, for $C$ classes. We do not employ directly the
joint probability $p(y_i,y_j)$ but we count conditionally and then average to
obtain a symmetric measure $p(y_i|y_j)$. Our final label compatibility is given
by the negative log-likelihood of this frequency:
\begin{equation}
\mu(y_i,y_j) = -\log\left(\frac{1}{2}\left(p(y_j|y_i) + p(y_i|y_j) \right)\right). \label{eq:mu}
\end{equation}
An example matrix for one of the dataset used in the experiments (Zeebrugges) is
illustrated in Figure~\ref{fig:coocc}. Note that we force a null diagonal to
enforce label smoothness.

\subsection{Inference in the Pylon model}

The Pylon model has been developed to perform semantic segmentation by jointly
inferring regions extent and labels issued from a tree representation. The model
imposes two main constraints on the segmentation: first, a single non-zero label
must be assigned to every pixel, what~\citet{lempitsky2011nips} call the
\emph{completeness constraint}. It implies that only one labeled region from the
tree can cover the pixel, while the other regions defined over a specific
location take the zero-label, i.e. inactive regions. This second requirement is
made explicit by the \emph{non-overlap constraint}, encoding the fact that
overlapping regions cannot take non-zero labels. Both constraints are trivially
met for a flat CRF, but become crucial constraints when making inference over a
hierarchy of overlapping segments. We refer the interested reader
to~\citep{lempitsky2011nips} for details about the reformulation of a CRF into a
Pylon model.

One of the most interesting aspects of the pylon model is that inference can be
performed by graph-cuts, thus taking advantage of the global optimality for the
two classes case and of the efficiency. In our setting, we employ
$\alpha$-expansion to perform inference with multiple classes. The Pylon models
reformulates inference as a linear program leading to a pseudo-boolean
optimization, which can be solved by the quadratic pseudo-boolean optimization
(QPBO)~\cite{rother2007cvpr}.

\section{Data and experimental setup}\label{sec:data-setup}

\subsection{Vaihingen benchmark}\label{sec:datava}

The Vaihingen dataset is a dataset provided by the International Society for
Photogrammetry and Remote Sensing (ISPRS), working group II/4, in
the framework of a ``2D semantic labeling contest'' benchmark.
\footnote{\url{http://www2.isprs.org/commissions/comm3/wg4/semantic-labeling.html}}.

The dataset is composed of 33 orthorectified image tiles acquired by a near
infrared (NIR) - green (G) - red (R) aerial camera, over the town of Vaihingen
(Germany). Images are accompanied by a digital surface model (DSM) representing
absolute height of pixels. \citet{gerke2015techrepo} released a normalized DSM,
which represents the pixels height relative to the elevation of the nearest
ground surface. We use this normalized DSM (nDSM) as an additional input to the
hypercolumn layer.

The average size of the tiles is $2'494 \times 2'064$ pixels with a spatial
resolution of 9~cm. 16 out of the 33 tiles are fully annotated at pixel level
and openly distributed to participants. In our experiments, we use 11 out of the
16 fully annotated image tiles for training (and hyperparameter selection),
while the remaining ones (tile ID 11, 15, 28, 30, 34) for validation, as
in~\citet{sherrah2016arxiv,volpi2017tgrs,maggiori2017tgrs}. The ground truths
maps for the remaining tiles are undisclosed and used to evaluate the accuracy
of submissions on an evaluation server run by the organizers.

The semantic segmentation task involves the discrimination of 6 land-cover /
land-use classification classes: ``impervious surfaces'' (IS) (roads, concrete
surfaces), ``buildings'' (BU), ``low vegetation'' (LV), ``trees'' (TR), ``cars''
(CA) and a class of ``clutter'' (CL) representing uncategorizable land covers.
For this problem, classes are highly imbalanced: classes ``buildings'' and
``impervious surfaces'' cover 50\% of the data, while ``car'' and
``clutter''only for 2\% of the total labels. Note that in our experiments,
following the evaluation practice, we do not predict the class ``clutter''. For
the color coding of this dataset, we refer to Table~\ref{tab:VaihingenSelf}.


\subsection{Zeebrugges benchmark}\label{sec:dataze}

This benchmark has been provided as part of the IEEE GRSS Data Fusion Contest in 2015 \citep{campos2016jstars} \footnote{\url{http://www.grss-ieee.org/community/technical-committees/data-fusion/2015-ieee-grss-data-fusion-contest/}}.

This dataset is composed of seven tiles of size $10'000 \times 10'000$ pixels.
Single images have a spatial-resolution of 5~cm, which we downgrade to 10~cm for
computational efficiency (in our past research~\citep{marcos2018jisprs} such
downgrade did not lead to losses in performances after final upsampling of the
predictions for evaluation). The images span RGB channels only. Five of the
seven images are released with labels~\citep{lagrange2015igarss} and used for
training and hyperparamter selection, while the remaining two are undisclosed
for evaluating the generalization accuracy on an evaluation server, according to
the challenge guidelines. Intuitively, the absence of a near infrared channel
makes this dataset very challenging, even after resampling. The Zeebrugges
dataset also comes with a LIDAR point cloud, that we transformed into a DSM by
gridding and averaging the point cloud. As the area is relatively flat, there is
no apparent need to normalize the DSM to relative heights. We include the DSM,
together with the RGB images, at the hypercolumn layer.

This problem involves 8 classes~\citep{campos2016jstars}: the same six as in the
Vaihingen benchmark, plus a land-cover ``water'' (WA) and a semantic class
``boats'' (BO). The class ``clutter'' (CL) represents a group of visual concepts
semantically more uniform than in the Vaihingen benchmark. This class groups now
mostly containers and other man made structures found in the harbor area. The
classes distribution is also very unbalanced: the  ``water'' class (in Vaihingen
it was part of the ``clutter'' class) composes almost 30\% of the annotated
data, while while ``cars'' and ``boats'' labels account for 1\% of the total.
The color coding is presented in~\ref{tab:ZeebruggesSelf}.

\subsection{Experimental Setup}

\paragraph{CNN architecture}

We first train the architecture presented in Section~\ref{sec:theory-cnn}. To
this end, we employ the standard VGG-16 network~\citep{simonyan2015iclr}, from
which we remove the last (5th) convolutional block and the fully connected
layers. Although the last layers could bring additional information, we argue
that they are too specialized on the ImageNet ILSVRC classes. The multi-task
layers learned on the hypercolumn stack have a similar purpose, which is to
learn a high-level, task specific, convolutional layers. To train the network,
we scale the learning rate of the pretrained VGG network to be 0.001 of the
global learning rate and we increase the weight decay by a factor 10. We set the
learning rate to $10^{-2}$ for Vaihingen and $10^{-3}$ for Zeebrugges. For
Vaihingen, we half its value every 100 epochs for a total of 300 epochs. The
weight decay has been set to 0.01 for 50 epochs to avoid initial spiking and
then to 0.0005 for the rest. Inputs are $256\times 256$ patches with a minibatch
size of 16. For Zeebrugges, we halve the learning rate after 200 epochs and
train for another 200 epochs, with a constant weight decay of 0.0001. For
Zeebrugges inputs are $500\times 500$ sub-tiles, and the minibatch set to 4. For
both cases, the loss is weighted by a inverse frequency strategy, truncated when
the inverse frequency is larger than 10. Note that a small random subset of the training patches is used for model selection, on which architectural choices and hyperparameters are selected. for both dataset, 10\% of the training patches are randomly held out at each run and used as a development set.

For the Vaihingen experiments, we learn 20-dimensional $3\times 3$ nonlinear
mappings from each VGG layer to the hypercolumn, with the exception of the last
layer which is only a $1\times 1$, thus being a fully connected layer. For the
Zeebrugges experiments, the architecture is the same but we employ
15-dimensional transitional layers. We fixed the kernel sizes to 3x3 and we
selected the dimensionality of the activation based on the error on a
small random held-out subset of the training set patches, after 50 epoch. Every
activation is upsampled at the original input size by using bilinear
interpolation, \emph{after} learning the layer specific convolutional mapping.
We did not experiment with learned upsampling and other complex interpolation
strategies. This line of works could be an interesting extension in the future.
We add two parallel $3 \times 3$ layers, mapping from the hypercolumn stack to a
25-dimensional activation space, followed by a fully connected layer for each
branch with the same dimensionality. One branch densely maps to semantic
classes, while the other scores densely the semantic boundaries.

\paragraph{Semantic boundary ground truths}

Our system is based on a two-tasks hypercolumn network, optimizing one loss on
the semantic segmentation task (Eq.~\eqref{eq:crossentropy}) and a second one on
the semantic boundaries extraction task (Eq.~\eqref{eq:Lb}). If the ground truth
for the semantic segmentation task is a classical ground truth as the one
provided in the two datasets used in this work (for an example in the Vaihingen
dataset, see Figure~\ref{fig:label}(b)), building a ground truth to learn the
semantic boundaries is less straightforward: to do so, we extracted all the
binary semantic boundaries from the ground truth (e.g. between buildings and the
rest, between cars and the rest) and used them as boundary classes (see
Figure~\ref{fig:edges} for an example of predictions reflecting this scheme). It
is built so that only 1 pixel thick line of pixels at the internal part of the
object are considered as edges, which means that two ground truth edges never
overlap and always lie one next to each other. Globally, for the employed
datasets, this means 5 GT for the Vaihingen and 8 GT for the Zeebrugges datasets.


\paragraph{Inputs}
As the input to VGG has to be three-dimensional, we input the nDSM only at the
hypercolumn layer. We believe this strategy is more natural: first, as
the VGG network is specialized in extracting visual concepts from natural
images, we let the pretrained network to work only on the optical domain, by
allowing some adjustment brought by fine tuning. This avoids an unnatural stack
of elevation and visual information directly at the input layer, or even more
artificial 3-dimensional inputs such as grayscale stacks. Although many
works underline that non-standard input spaces still perform well, e.g.
\citep{marmanis2018jisprs,sherrah2016arxiv}, we prefer to use nDSM as a feature
and not as a visual input. Secondly, by following the hypercolumn philosophy,
VGG can be seen as a feature extractor and only after we learn a ``second''
convolutional neural network performing segmentation. Summing up, at the
hypercolumn layer we concatenate i) elevation information, ii) the original
input image and iii) the activations learned from the VGG. Studying the effects
of concatenating features learned by specific networks at the hypercolumn stack
could be another interesting line of future research. In this work, we
rely on ideas presented in \citep{maggiori2017tgrs,marcos2018jisprs} where
networks perform first feature extraction and, from the hypercolumn, learn
semantic segmentation. Following these intuitions, we let the network learn to
combine features, from both the network activation and the visual and elevation
input domains.

\paragraph{CRF and Pylon}
We tune the hyperparameters of the segmentation tree and of the CRF using a
random subset of the training patches. Note that this step follows the CNN
training and the computation of the superpixels. Specifically, once the CNN is
trained, we estimate superpixels based on \citep{dollar2013iccv}, which links
predicted edge strength with color cues. Then, we estimate the UCM by the
approach and toolbox of \citep{arbelaez2011tpami}. To generate the segmentation
tree, we estimate the hyperparameters of the dissimilarity metric presented in
Section~\ref{sec:theory-ucm} by measuring on training images how the generated
regions match their median label at different levels of the tree. We also
adjusted some hyperparameters controlling the region size and edge-versus-color
regularization.

Regarding the CRF potentials, we learn co-occurrence scaling
($\cdot\mu(y_i,y_j)$ in Eq.~\eqref{eq:mu}) by counting first order interactions
among the labels observed at the superpixels (leaf) level in the training set.
We simply count frequencies of classes appearing on two neighboring superpixels
over all the available training images. We assume that this distribution is the same as the one of test images, which is often true. The other parameters of the CRF are tested by measuring the segmentation accuracy on the random validation subset of patches from the training set.

\section{Results}\label{sec:results}

\paragraph{Baselines} We compare the proposed segmentation pipeline to different
baselines. The first, named \textbf{Unary PX}, evaluates the segmentation
accuracy as given by the pixelwise prediction from the CNN. The second is named
\textbf{Unary SP} and reports the accuracy of superpixels labeling. Pixel-based
likelihoods are averaged for each superpixel and the maximum-a-posteriori for
each region provides the final labels. Note that superpixels represent the
lowest level of the segmentation tree and these are produced by aggregating edge
predictions for each class and performing a watershed oversegmentation
\citep{dollar2013iccv}. The third and fourth baselines exploit a CRF on top of
the \textbf{Unary SP} aggregation. The first CRF  model \textbf{CRF Color} --
implements a standard CRF, using a contrast sensitive potential applying
Eq.~\eqref{eq:phih} and using pairwise co-occurrences. This implement as simple
yet effective baseline. The second CRF uses all the potentials described in the
paragraph ``interaction terms'' in Sec.~\ref{sec:crf}, in a flat graph
structure. We name it \textbf{CRF Flat}. Finally, the method making use of the
segmentation tree is named \textbf{CRF Tree}. The aim of these comparison is to
show the improvement brought by every module in the pipeline.

\paragraph{Accuracy metrics} We report the overall accuracy (OA, total number of
correctly predicted pixels over total number of labeled pixels), the average
accuracy (AA, or recall, fraction of correctly retrieved pixels for each
class, averaged) and the F1-score (F1, geometric mean between precision and
recall, averaged for each class). These measures show global accuracy
independently of class size (OA), global accuracy by considering all classes
equally important (AA) and global accuracy by considering over- and
under-predictions for each class (F1). We also report the per-class F1
score for the teste baselines and proposed pipeline. Finally, we shortly
report and discuss results from other papers working on the same datasets.

Furthermore, we measure the accuracy of the predicted semantic edges by
plotting ROC curves and comparing the area under the curve (AUC). To do so, we
use the raw edge likelihood as provided by the network, after a step of
non-maximum suppression. We use two different edge ground truths for the
evaluation: the first one has a 1-pixel edge line, while the second a 3-pixel
width one. Although the network has been trained with the former one (also see
Figure~\ref{fig:edges}), we evaluate also with the latter, which is more
permissive. We do that because the ground truths for both datasets have been
manually generated by photointerpretation and are thus far from being accurate
on objects borders, a fact also worsened by orthorectification artifacts.

\subsection{Vaihingen}

\definecolor{IS}{rgb}{1,1,1}
\definecolor{WA}{rgb}{0,0,0.5}
\definecolor{CL}{rgb}{1,0,0}
\definecolor{LV}{rgb}{0,1,1}
\definecolor{BU}{rgb}{0,0,1}
\definecolor{TR}{rgb}{0,1,0}
\definecolor{BO}{rgb}{1,0,1}
\definecolor{CA}{rgb}{0.5,0.2,1}

\fboxrule=0.3mm
\fboxsep=0mm
\newcommand\csqua[1]{\fbox{{\textcolor{#1}{\rule{4mm}{2mm}}}}}

\renewcommand{\tabcolsep}{2pt}
\begin{table}[!t] \centering
\small{
\begin{tabular}{r|ccccc|ccc}
  \toprule
  {\bf Color code} & \csqua{IS} & \csqua{BU} & \csqua{LV} & \csqua{TR} & \csqua{CA}   & & & \\
{\bf Model} & IS & BU & LV & TR & CA & {\bf OA} & {\bf AA} & {\bf F1} \\ \hline
{\bf Unary PX}  & 85.36 & 90.97 & 72.05 & 84.19 & 69.70 & 83.38 & {\bf 82.75} & 80.45 \\
{\bf Unary SP}  & 86.57 & 91.68 & 73.23 & {\bf 84.83} & 71.00 & 84.33 & 82.33 & 81.46 \\
{\bf CRF Color} & 86.10 & 91.41 & 72.43 & 83.18 & 71.54 & 83.55 & 80.68 & 80.93 \\
{\bf CRF Flat} 	& 85.72 & 91.48 & 72.21 & 82.83 & 71.76 & 83.31 & 80.01 & 80.80 \\
{\bf CRF Tree} 	& {\bf 86.80} & {\bf 91.85} & {\bf 73.80} & 84.57 & {\bf 73.82} & {\bf 84.50} & 82.16 & {\bf 82.17} \\
\bottomrule
\end{tabular}}
\caption{Semantic segmentation accuracies on the Vaihingen validation  set. F1 scores evaluate per-class accuracy. Bold denotes best results. Models not employing CRF tend to provide more accurate segmentation of small classes (e.g. cars), as indicated by the highest AA. CRF-based methods helps in balancing errors on both small and large classes, achievingbest average F1 scores.}
\label{tab:VaihingenSelf}
\end{table}

\renewcommand{\tabcolsep}{3pt}
\begin{figure}[!t] \centering
\begin{tabular}{cc}
{\bf AUC 1px}&{\bf AUC 3px}\\
{\includegraphics[width=0.4\textwidth]{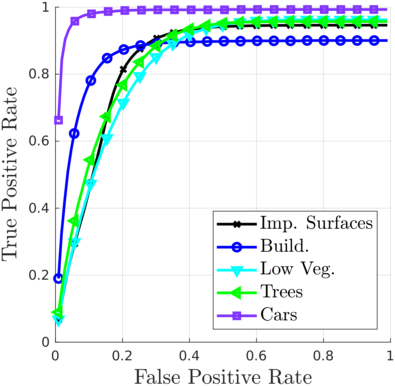}}&                     {\includegraphics[width=0.4\textwidth]{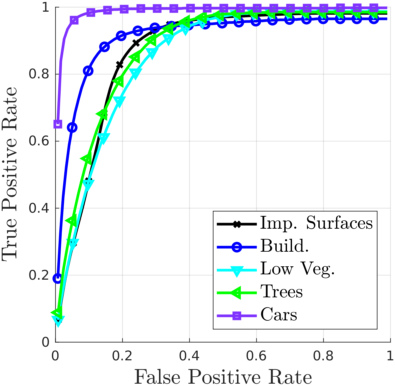}}\\
\end{tabular} \small{
\begin{tabular}{c|ccccc|c}
\toprule {\bf Class} & IS    & BU    & LV    & TR    & CA & Mean \\ \hline {\bf AUC 1px} & 83.53 & 85.53 & 82.59 & 84.37 & 97.64 & 86.73 \\ 
{\bf AUC 3px} & 86.58 & 90.80 & 85.10 & 87.17 & 98.23 & 89.58 \\ \bottomrule 
\end{tabular}
}
\caption{Average per-class ROC curves (across validation images) and
average per-class AUC on the Vaihingen validation set. Prediction of the 1-pixel
wide edges is a challenging problem. 3-pixel wide edges are much more accurate,
in particular for spatially smooth classes. The creation of the UCM allows to
cope with blurred predictions.}
\label{fig:VaihingenAUC}
\end{figure}

\newcommand{\sz}{0.2}

\renewcommand{\tabcolsep}{1pt}
\begin{figure*}[!t]
\centering
\includegraphics[width=.8\textwidth]{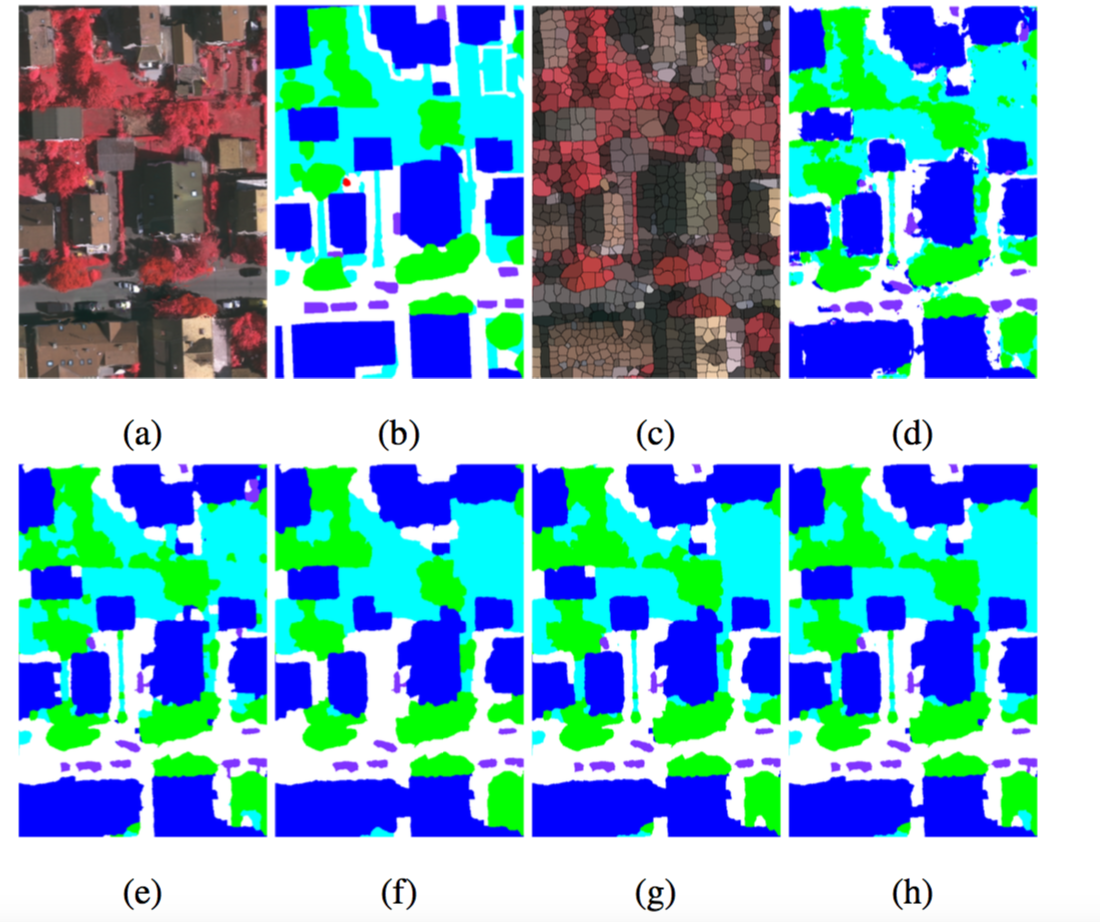}
\caption{Examples of semantic segmentation on zoom of a validation image. (a) original image, (b) ground truth, (c) Superpixels, (d) Unary PX, (e) Unary SP, (f) CRF Color, (g) CRF Flat, (h) CRF Tree. Oversmoothing on cars (purple) is visible. However, larger and smoother structures are better delineated. \textbf{(Note: in this preprint we had to decrease the graphics resolution. For full resolution, please refer to the published version, or contact the authors)}}
\label{fig:label}
\end{figure*}

Table~\ref{tab:VaihingenSelf} shows numerical results obtained on the Vaihingen
validation set. The proposed pipeline achieves the best overall accuracy and
best F1 score compared to competitors. {\bf Unary PX} and {\bf Unary SP}
show slightly better AA (+0.59 and +0.07, respectively) thanks to an inferior
oversmoothing of small classes. The F1 scores indicate that {\bf CRF Flat} and
{\bf CRF Tree} achieve more balanced average precision and average recall
scores. Combining this observation with the fact that CRFs tend to oversmooth
small classes (mostly ``car'') with the F1 scores, it seems that Unary methods
also tend to overpredict them, making {\bf CRF Tree} achieve more balanced
segmentations, i.e. better F1 scores.

Figure~\ref{fig:label} details some examples of semantic segmentation results.
It only focuses on a detail of a validation map (ID 28, upper left corner).
Figure~\ref{fig:label}(c) shows the common superpixel representation obtained by
the edge pooling step, which is then used to group probabilities from {\bf Unary
PX} into {\bf Unary SP} maps (Figure~\ref{fig:label}(d) and
Figure~\ref{fig:label}(e), respectively). Spatially pooling predictions in
superpixels remove spurious noise and spatially smoothes class labels, as
expected, which is reflected by improved or worsened class-specific accuracies,
depending on the degree of smoothing beneficial to such class. Even without
explicitly modeling context, maps are highly improved visually. As superpixels
are generated accordingly to learned semantic boundaries, spatial pooling
respects actual class borders more rather than pure color homogeneity of
superpixels, as commonly implemented.

Maps generated by CRF models (Figure~\ref{fig:label}(f)-(h)) tend to switch
labels for regions which are ambiguously assigned to a class. We typically
observe improved segmentation across objects edges, where the {\bf Unary PX}
model provides spatially inhomogeneous labelings. These improvements are
expected when modeling context and spatial smoothness by a CRF. The {\bf CRF
Tree} model (Figure~\ref{fig:label}(h) finds a good compromise between the
oversmoothing offered by the {\bf CRF Color} (Figure~\ref{fig:label}(f)) and the
preservation of detail given by {\bf Unary SP} Figure~\ref{fig:label}(e). {\bf
CRF Flat} seems to perform in between the two, as one would expect.

Prediction of semantic edges is accurate overall, with the tendency of being
more accurate for well-defined objects (cars, buildings, trees) rather than for
amorphous classes. The increase in the AUC scores from 1 pixel (1px) to 3 pixels
(3px) indicates that (assuming the ground truths are precise enough) the
prediction is dilated for classes that have mostly ambiguous boundaries, complex
shape or wide and smooth transitions between colors and labels.

\renewcommand{\tabcolsep}{2pt}
\begin{table}[!t] \centering
\small{
\begin{tabular}{r|ccccc|ccc}
  \toprule
  {\bf Reference} & {\bf OA} & {\bf AA}  \\ \hline
\citep{marmanis2018jisprs}, boundary, single scale  & 84.8 &  - \\
\citep{marmanis2018jisprs}, no boundary, ensemble   & 85.5 &  - \\
\citep{marmanis2018jisprs}, full model              & 89.8 &  - \\
\citep{marcos2018jisprs}, rotation equivariant network                                             & 87.5 & 83.9 \\
\citep{volpi2017tgrs}, SegNet                       & 87.8 & 81.3 \\
\citep{audebert2018jisprs}, SegNet                  & 89.4 & - \\
Tang, W. (Abbrev.: WUH$\_$W3) $\dagger$, ResNet-101, \`a-trous conv.
                                                    & 89.7 & - \\
Li, H. (Abbrev.: CAS$\_$L1)$\dagger$, PSPNet        & 85.7 & - \\
Sun, Y. (Abbrev.: HUSTW5)$\ddagger$ Ensemble of Deconv. Net and U-Net                                               & 91.6 \\
\hline
Ours                                                & 84.50 & 82.16 \\
\citet{marcos2018jisprs}, baseline                  & 87.4 & 78.2 \\
\bottomrule
\end{tabular}}
\caption{Comparison to state-of-the-art approaches from the literature on the ISPRS Vaihingen dataset. $\dagger$ and $\ddagger$ are taken from the ISPRS Vaihingen challenge website ranking (\url{http://www2.isprs.org/commissions/comm2/wg4/vaihingen-2d-semantic-labeling-contest.html}) as representative of recent computer vision methods, and $\ddagger$ is the current best performing method on the challenge, as per May 8, 2018. \citep{marcos2018jisprs} baselines denotes a CNN with a number of parameters close to our VGG-derived trunk.}
\label{tab:Vaihingensota}
\end{table}

Table~\ref{tab:Vaihingensota} shows comparison to state-of-the-art
approaches. We list approaches covering different families of architectures,
which in turn also provide very diverse accuracies. The best performing
strategies rely on ensembling of the prediction of multiple networks, while very
deep and complex segmentation approaches do improve the results, but not
significantly over fully convolutional networks and standard SegNets. Our
approach compares similarly to other standard fully convolutional networks.
Differences are mostly visual, where our approach provides an improved spatial
regularization at the cost of some oversmoothing. However, the main trunk we use
(VGG) can replaced by any modern architecture in a straightforward way, and lead
to improved performances without any modification to the method itself.


\subsection{Zeebrugges}

\renewcommand{\tabcolsep}{2pt}
\begin{table}[!t]
  \centering
\scriptsize{
\begin{tabular}{r|cccccccc|ccc}
  \toprule
{\bf Color code} & \csqua{IS} & \csqua{WA} & \csqua{CL} & \csqua{LV} & \csqua{BU} & \csqua{TR} & \csqua{BO} & \csqua{CA} & & & \\
{\bf Model}    & IS & WA & CL & LV & BU & TR & BO & CA & {\bf OA} & {\bf AA} & {\bf                                                                   F1} \\ \hline
{\bf Unary PX} 	& 80.98 & 98.62 & 61.41 & 80.31 & 80.60 & 57.71 & 57.77 & {\bf 78.12} & 84.51 & 74.47 & 74.44 \\
{\bf Unary SP}  & 82.48 & 98.88 & 66.06 & 80.99 & 82.10 & {\bf 59.24} & 62.47 & 77.47 & 85.54 & {\bf 75.81} & 76.21 \\
{\bf CRF Color} & 83.08 & 98.97 & 67.55 & {\bf 81.47} & 82.82 & 56.43 & 63.78 & 72.53 & 85.89 & 74.52 & 75.83 \\
{\bf CRF Flat} 	& {\bf 83.94} & 98.81 & 67.24 & 80.88 & {\bf 84.48} & 48.01 & 60.51 & 72.03 & 86.00 & 71.24 & 74.49 \\
{\bf CRF Tree} 	& 83.15 & {\bf 98.94} & {\bf 68.12} & 81.21 & 83.44 & 58.34 & {\bf 65.65} & 76.96 & {\bf 86.05} & 75.09 & {\bf 76.98} \\
\bottomrule
\end{tabular}
}
\caption{Semantic segmentation accuracies on the Vaihingen validation  set. F1 scores evaluate per-class accuracy. Bold denotes the best results. Models not employing CRF do not oversmooth spatially small classes (e.g. cars) by a large margin, while CRF helps in removing errors in spatially smooth structures. {\bf CRF Tree} finds a good balance between these solutions.}
\label{tab:ZeebruggesSelf}
\end{table}

\renewcommand{\tabcolsep}{2pt}
\begin{figure}
\centering
\begin{tabular}{cc}
{\bf AUC 1px}&{\bf AUC 3px}\\
{\includegraphics[width=0.4\textwidth]{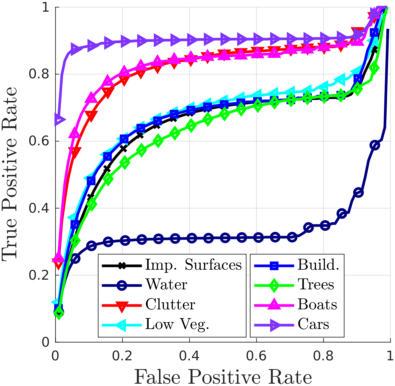}}&
{\includegraphics[width=0.4\textwidth]{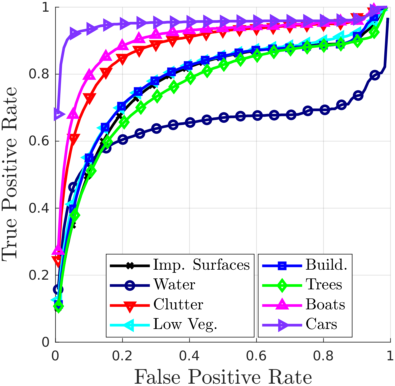}}\\
\end{tabular}
\scriptsize{
\begin{tabular}{c|cccccccc|c}
\toprule
{\bf Class} & IS & WA & CL & LV & BU & TR & BO & CA & Mean \\ \hline
{\bf AUC 1px} & 65.15 & 33.60 & 81.75 & 67.90 & 66.66 & 62.96 &82.41 & 89.63 & 68.76 \\
{\bf AUC 3px} & 77.67 & 64.50 & 87.80 & 79.05 & 78.60 & 75.78 &89.75 & 94.64 &  80.97 \\ \bottomrule
\end{tabular}
}
\caption{Average per-class ROC curves and average per-class AUC on the
Zeebrugges test set. Edge probabilities are less sharp than the Vaihingen ones
(see sharp increase for the large false-positive rates), suggesting that the
problem might be harder. This is also supported by the larger gap between 1- and
3-px predictions.}
\label{fig:ZeebruggesAUC}
\end{figure}

\renewcommand{\tabcolsep}{1pt}
\begin{figure*}
\centering
\includegraphics[width=.9\textwidth]{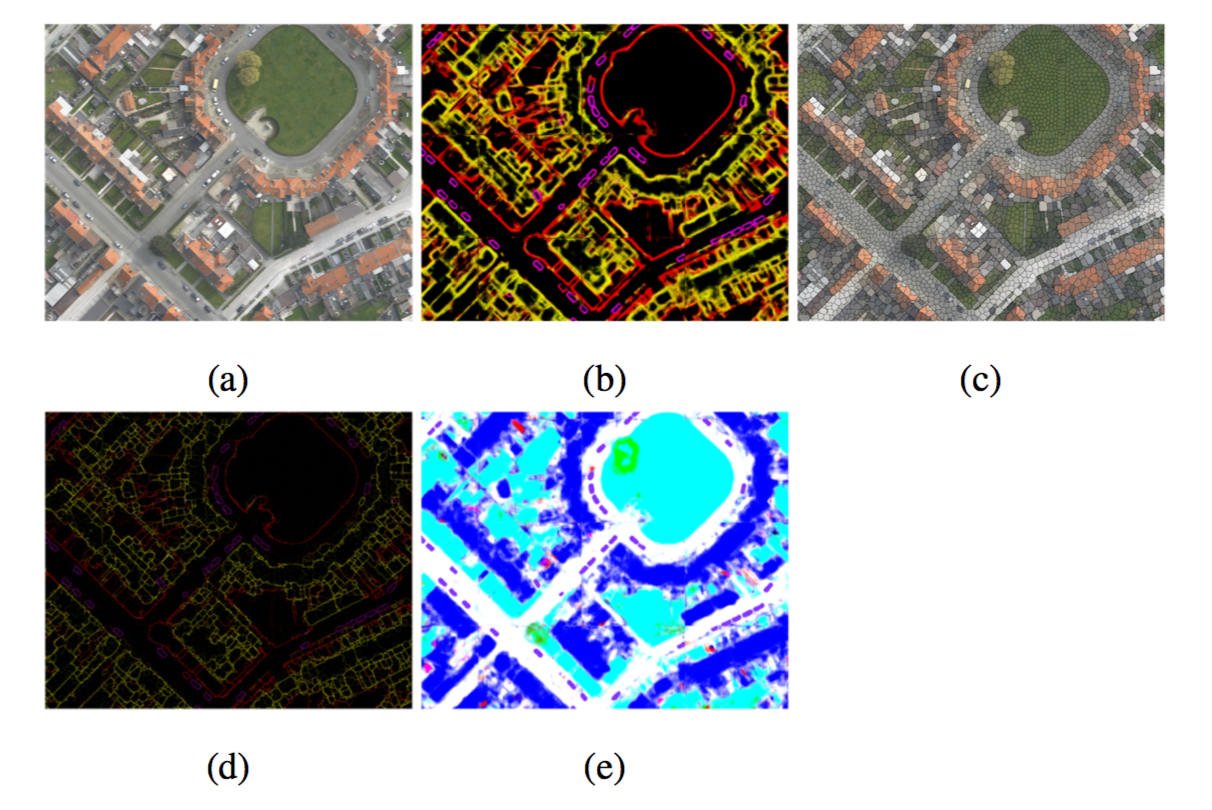}
\caption{Example of semantic boundary detection on a test image zoom-in.
(a) original image, (b) predicted edges, (c) superpixels,  (d) UCM, (e) estimated class likelihoods. \textbf{(Note: in this preprint we had to decrease the graphics resolution. For full resolution, please refer to the published version, or contact the authors)}}
\label{fig:edges}
\end{figure*}

The results on the Zeebrugges dataset in Table~\ref{tab:ZeebruggesSelf} reflect
similar observations, with {\bf CRF Color} performing slightly better on
average. {\bf CRF Tree} and {\bf CRF Flat} provide the highest OA. In terms of
AA, {\bf Unary SP} still provides best numbers, since it avoids oversmoothing.
However, in terms of balance between commission and omission errors (F1 score),
{\bf CRF Tree} outperforms all competitors, +0.77 F1 points.

Figure~\ref{fig:edges} presents a detail of the results obtained on one of the
two test images. In Figure~\ref{fig:edges}(b), we show edge likelihoods for
three classes in a RGB composition: ``impervious surfaces'', ``buildings'' and
``cars''. On screen, the edges overlap: while the contour color for ``impervious
surfaces'' is red, buildings are almost always contoured also by the
aforementioned class, making their contour yellow (red and green). For the same
reason, cars appear in purple, a composition of red and blue. Note that a given
pixel can have high likelihood for more than one semantic boundary class.

It is interesting to note that the class ``buildings'' shows high internal
variability, since the class is cluttered with edges and boundaries that are
similar to actual semantic boundaries. Superpixels shown in
Figure~\ref{fig:edges}(c) are obtained by employing predicted edges into a
watershed transform. It clearly appears that the semantic edges are more exactly
followed than other boundaries, making the {\bf Unary PX} pooling into {\bf
Unary SP} more accurate. Merging predicted edges likelihoods and superpixels
edges into the UCM map (Figure~\ref{fig:edges}(d)) has the effect of thinning
the predicted likelihoods and making them form a segmentation tree for each
class. Figure~\ref{fig:edges}(e) shows the raw class likelihoods ({\bf Unary
PX}) the network predicts for the same area. The more saturate the color is, the
more confident is the prediction.

Figure~\ref{fig:fullZElabel} shows full predictions maps for the Zeebrugges test
tile representing the town center. Different smoothing levels can be appreciated
in particular in large semantically homogeneous regions and around object
borders.

Contrarily to the Vaihingen dataset, 1px edges are not predicted well, overall.
This is due to to mostly imprecise delineation of the ground truth. Although the
CNN is able to learn where in a given field-of-view an edge should likely
appear, at test time this results in rather smooth and fuzzy edges. Also, the
variance of flat uniform surfaces (e.g. ``Water'') makes the prediction of
actual semantic boundaries fuzzy, an effect quantified by the large gap between
1px and 3px AUC. As for the Vaihingen benchmark, classes with relatively
straight and sharp boundaries (such as ``clutter'', ``boats'' or ``cars'') are
the ones delineated best. Figure~\ref{fig:fullZEedges} shows image-scale
predictions of semantic edges, for each class independently.

\renewcommand{\tabcolsep}{2pt}
\begin{table}[!t] \centering
\small{
\begin{tabular}{r|ccccc|ccc}
  \toprule
  {\bf Reference} & {\bf OA} & {\bf AA}  \\ \hline
\citep{campos2016jstars}, VGG/SVM                   & 76.6 & - \\
\citep{campos2016jstars}, AlexNet                   & 83.3 & - \\
\citep{marcos2018jisprs}, rotation equivariant network                                             & 82.6 & 75.3 \\
RIT1$\ddagger$, [method not reported] & 87.9 & - \\
\hline
Ours                                                & 86.0 & 75.1  \\
\bottomrule
\end{tabular}}
\caption{Comparison to state-of-the-art results on the Zeebrugges challenge dataset. $\ddagger$ is the current best performing method on the challenge leaderboard, as per May 8, 2018.}
\label{tab:Zeebruggessota}
\end{table}
Table~\ref{tab:Zeebruggessota} shows a comparison to available
state-of-the-art results on the Zeebrugges dataset. Other approaches rely on
relatively standard architectures and schemes, which are outperformed by our
regularization scheme. The only notable exception is the current leader of the
challenge (RIT1), but no details about the method used are currently reported.
However, we note that this dataset has received less attention than the
Vaihingen benchmark.


\section{Discussion}\label{sec:discussion}

The basic {\bf Unary PX} and {\bf Unary SP} perform really well in terms
of AA. The reason is that they do not tend to oversmooth predictions and, even
when {\bf Unary SP} does, it oversmooths only locally in the extent of the
superpixels. Since the shape of superpixels directly follows semantic boundaries
in the images, average class-likelihoods in regions tend to be positively
correlated with the actual label, filtering out spurious noise. Overmoothing
concerns mostly spatially small classes, where CRFs tend to favor the class
surrounding locally the smaller ones, missing some pixels of such objects.
However, modeling spatial interactions allows to remove erroneous localizations,
which is reflected by better F1 scores. For instance, accounting for
relationships of class ``cars'' to other classes in the output space, allows to
remove erroneous occurrences, such as mislabeling of car-like structures on top
of buildings.

CRF are the best options. In the proposed pipeline and most baselines, CRFs are
not only smoothing schemes relating on local likelihoods and color distribution
but they take into account the spatial arrangement of classes (co-occurrences)
together with color, deep features, relative elevation and semantic edges
between them. This makes the smoothing scheme better reflect the prior
information about the problem, as well as the prior knowledge acquired from
training data (or possibly unrelated ancillary sources, when available). In
addition, exploiting a hierarchical representation of regions further helps in
smoothing only selectively, where accommodated by the energy minimization
process.

The simultaneous prediction of semantic edges by the multi-task CNN seems
accurate, but obviously highly correlated to the quality (and availability) of a
dense ground truth used for learning. During training, the CNN seems robust to
some errors in edge localization: the translation invariance property of the VGG
base trunk allows the hypercolumn to contain all the relevant information to
compensate for systematic shifts and biases in the ground truth. The differences
between the 1px and 3px semantic edge evaluations account for these effects. In
any case, edge thinning and postprocessing into superpixel edges allows for an
accurate and informative UCM, impacting positively the resulting segmentation
tree and the semantic segmentation overall.

Furthermore, the presented pipeline allows to use any available CNN
network as the main trunk in the feature extraction network, providing the
hypercolumn features. We argue that more accurate networks would provide more
expressive features, leading to better segmentation quality overall. On another
line, if the ground truth is limited, other schemes such as the one presented in
\citep{marcos2018jisprs} can be adopted. We posit that the presented
architecture is generic and allows for data and problem specific regularization,
\emph{on top} of an already trained network. If the data coming from the domain
specific problem (e.g. the Zeebrugges dataset) is not enough to train or
properly fine tune a very deep network, it might be enough to learn multi-task
layers modeling specific features of the problem with different sources of data,
on top of the available architectures trained on natural images.



It is worth mentioning that we did not spend significant amount of time in fine
tuning the CNN architecture and hyperparamters, nor the CRF potentials, so we
assume that further improvements can be obtained by extended hyperparameters and
architecture search.

\renewcommand{\sz}{0.25}

\begin{figure}[!th]
\renewcommand{\tabcolsep}{1pt}
\centering
\includegraphics[width=.9\textwidth]{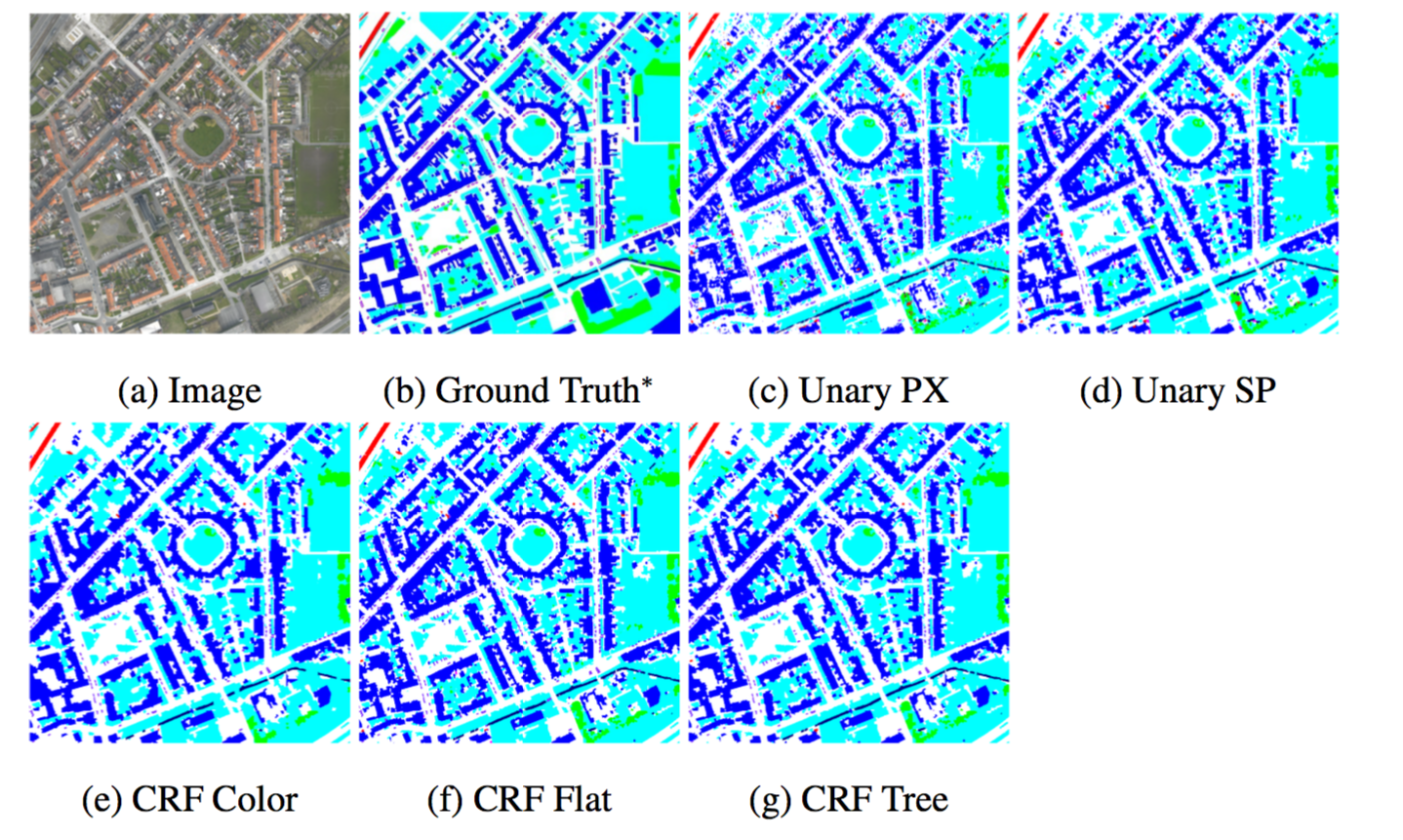}
\caption{Example of semantic segmentation on a full Zeebrugges test image. (a) original image, (b) ground truth$^*$, (c) Unary PX, (d) Unary SP, (e) CRF Color, (f) CRF Flat, (g) CRF Tree. Better viewed in the electronic version. Overall, CRF offer smoother and maps which are easier to read. Unary-derived maps are much more noisy and cluttered. ${}^*$The ground truth has been screen-printed from \citep{lagrange2015igarss} and color for the ``car'' ground truth modified in a photo-editing software, resulting slightly blurred. \textbf{(Note: in this preprint we had to decrease the graphics resolution. For full resolution, please refer to the published version, or contact the authors)}}
\label{fig:fullZElabel}
\end{figure}

\begin{figure}[!th]
\renewcommand{\tabcolsep}{1pt}
\centering
\includegraphics[width=.9\textwidth]{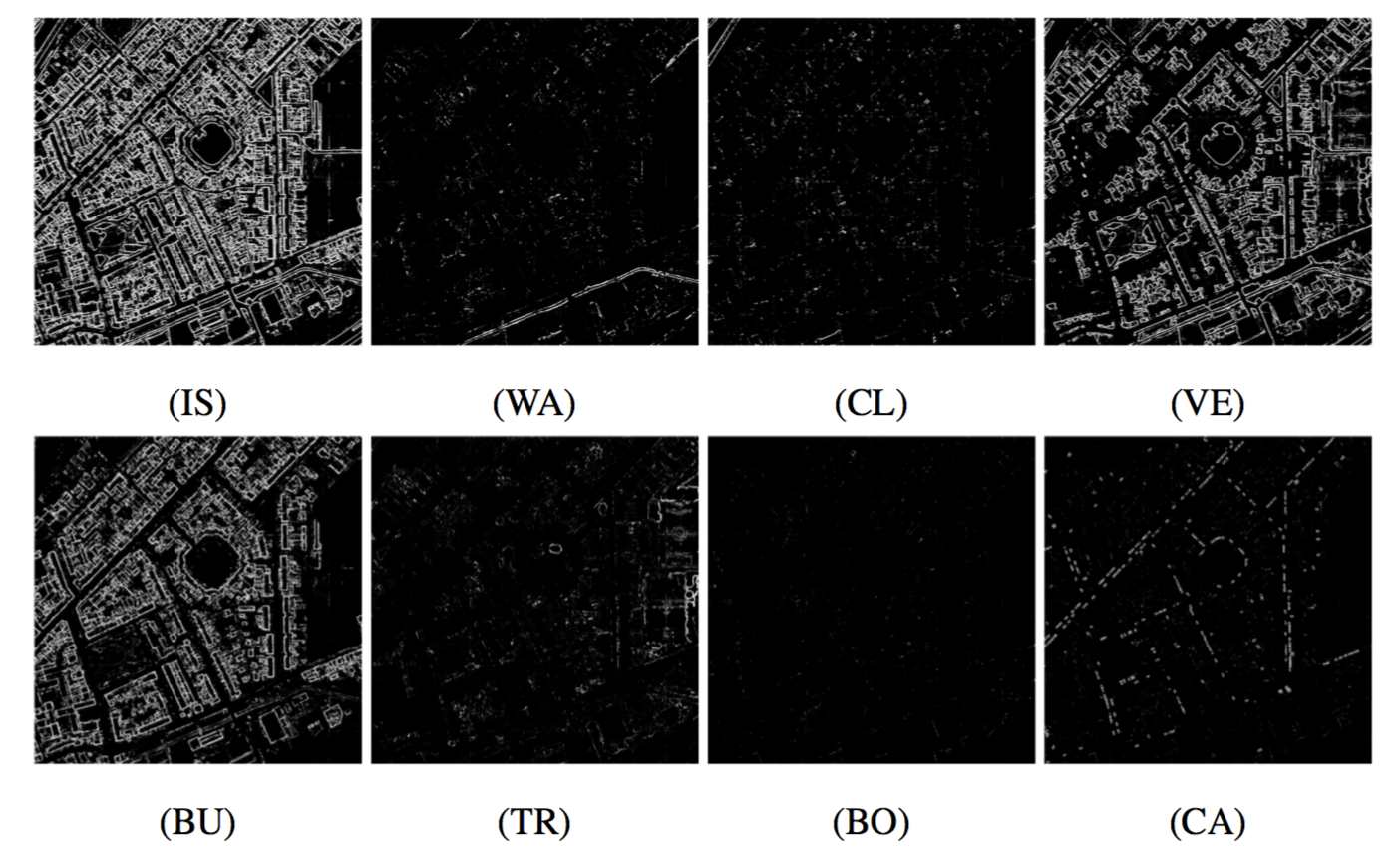}
\caption{Per-class semantic edges on the same Zeebrugges image. Better viewed in the electronic version. Overall, structures are well delineated indicating that learning semantic boundaries in overhead imagery leads semantically coherent results \textbf{(Note: in this preprint we had to decrease the graphics resolution. For full resolution, please refer to the published version, or contact the authors)}.}
\label{fig:fullZEedges}
\end{figure}

\section{Conclusions and future perspectives}\label{sec:concs}

We proposed a model fusing bottom-up hierarchical evidence about local
appearance and top-down prior information about local spatial organization and
pairwise relationships between superpixels.

Regarding the first aspect, we showcased the possibility of learning the tasks
of dense semantic segmentation and semantic boundaries \emph{jointly}, in an
end-to-end way, using a modified pretrained network. To do so, we extended a CNN
formulation for multi-task learning, relying on the hypercolumn architectures
built around a truncated pretrained VGG network. In our setting, we proposed to
control the number of learnable parameters by forcing intermediate layers
stacking into the hypercolumn to act as information bottlenecks distilling
information from the original VGG activations. It results that the hypercolumn
stack contains only the relevant information to solve the tasks at hand, which
we assume to be less complex than vision tasks such as the ILSVRC ImageNet
classification task. As the last convolutional block and the fully connected
layers of the original VGG network are not dropped, the size of the network we
adopted in this paper is still smaller that the full VGG. Further
improvement might be gained by employing other solutions in terms of modern CNN
architectures.

Regarding the second aspect, we built a CRF model specifically tailored to
segment color and elevation information, with pairwise potentials accounting for
the expected aspects of both data sources. We also made use of statistics
collected on the training data to facilitate or to avoid specific class
combinations. Thanks to the dense semantic edge likelihood, we were able to
translate this information first into superpixels and then unto a segmentation
tree by means of the UCM representation. We performed inference by employing the
Pylon model wrapping a graph-cut solver (QPBO). Both the numerical accuracy and
the overall readability of the maps were improved by injecting prior knowledge
about typical layout configurations.

Future research will explore how to further simplify the segmentation map while
preserving geometrical features, in order to achieve accurate vectorization and
generalization without losing important details. It would be interesting to do
so by specifying a generalization level e.g. corresponding to probable cuts in
the segmentation tree. Another interesting research direction would be to learn
pairwise potentials used in this work directly from data, since some class
combinations would require more or less influence from specific potentials. This
combination could be learned using structured learning, similarly to
~\citep{volpi2015bcvprw}.

\section*{Acknowledgments}

This work was partly supported by the Swiss National Science Foundation, grant
150593 ``Multimodal machine learning for remote sensing information fusion''
(http://p3.snf.ch/project-150593). The authors would also like to thank the
Belgian Royal Military Academy, for acquiring and providing the Zeebrugges data
used in this study, ONERA (The French Aerospace Lab), for providing the
corresponding ground-truth data~\cite{lagrange2015igarss}, and the IEEE GRSS
Image Analysis and Data Fusion Technical Committee for running the custom
evaluation for the semantic boundary detection task.

\section{References}

\bibliographystyle{elsarticle-harv}
\biboptions{authoryear}
\bibliography{regParsing-jisprs}

\begin{thebibliography}{47}
\expandafter\ifx\csname natexlab\endcsname\relax\def\natexlab#1{#1}\fi
\expandafter\ifx\csname url\endcsname\relax
  \def\url#1{\texttt{#1}}\fi
\expandafter\ifx\csname urlprefix\endcsname\relax\def\urlprefix{URL }\fi

\bibitem[{Arbelaez et~al.(2011)Arbelaez, Maire, Fowlkes, and
  Malik}]{arbelaez2011tpami}
Arbelaez, P., Maire, M., Fowlkes, C., Malik, K., 2011. Contour detection and
  hierarchical image segmentation. IEEE TPAMI 33~(5), 898--916.

\bibitem[{Asner et~al.(2005)Asner, Knapp, Broadbent, Oliveira, Keller, and
  Silva}]{asner2005science}
Asner, G.~P., Knapp, D.~E., Broadbent, E.~N., Oliveira, P. J.~C., Keller, M.,
  Silva, J.~N., 2005. Selective logging in the brazilian amazon. Science 310,
  480.

\bibitem[{Audebert et~al.(2018)Audebert, Le~Saux, and
  Lef\`evre}]{audebert2018jisprs}
Audebert, N., Le~Saux, B., Lef\`evre, S., 2018. Beyond {RGB}: Very high
  resolution urban remote sensing with multimodal deep networks. ISPRS Journal
  of Photogrammetry and Remote Sensing 140, 20--32.

\bibitem[{Besag(1974)}]{besag1974jrss}
Besag, J., 1974. Spatial interaction and the statistical analysis of lattice
  systems. Journal of the Royal Statistical Society. Series B 36~(2), 192--236.

\bibitem[{Boykov et~al.(2001)Boykov, Veksler, and Zabih}]{boykov2001tpami}
Boykov, Y., Veksler, O., Zabih, R., 2001. Fast approximate energy minimization
  via graph cuts. IEEE Transactions on Pattern Analysis and Machine
  Intelligence 23~(11), 1222--1239.

\bibitem[{Campbell et~al.(1997)Campbell, Mackeown, Thomas, and
  Troscianko}]{campbell1997pattrec}
Campbell, M.~W., Mackeown, W. P.~J., Thomas, B., Troscianko, T., 1997.
  Interpreting image databases by region classification. Pattern Recognition
  30~(4), 555 -- 563.

\bibitem[{Campos-Taberner et~al.(2016)Campos-Taberner, Romero-Soriano, Gatta,
  Camps-Valls, Lagrange, Saux, Beaup{\`e}re, Boulch, Chan-Hon-Tong, Herbin,
  Randrianarivo, Ferecatu, Shimoni, Moser, and Tuia}]{campos2016jstars}
Campos-Taberner, M., Romero-Soriano, A., Gatta, C., Camps-Valls, G., Lagrange,
  A., Saux, B.~L., Beaup{\`e}re, A., Boulch, A., Chan-Hon-Tong, A., Herbin, S.,
  Randrianarivo, H., Ferecatu, M., Shimoni, M., Moser, G., Tuia, D., 2016.
  Processing of extremely high resolution {LiDAR} and {RGB} data: Outcome of
  the {2015 IEEE GRSS Data Fusion Contest. Part A: 2D} contest. IEEE J. Sel.
  Topics Appl. Earth Observ. Remote Sens. 9~(12), 5547--5559.

\bibitem[{Crommelinck et~al.(2016)Crommelinck, Bennett, Gerke, Nex, Yang, and
  Vosselman}]{crommelinck2016remsens}
Crommelinck, S., Bennett, R., Gerke, M., Nex, F., Yang, M.~Y., Vosselman, G.,
  2016. Review of automatic feature extraction from high-resolution optical
  sensor data for uav-based cadastral mapping. Remote Sensing 8~(8).

\bibitem[{Dollar and Zitnick(2013)}]{dollar2013iccv}
Dollar, P., Zitnick, C., 2013. Structured forests for fast edge detection. In:
  International Conference on Computer Vision (ICCV).

\bibitem[{Felzenszwalb and Huttenlocher(2004)}]{felzenszwalb2004ijcv}
Felzenszwalb, P., Huttenlocher, D., 2004. Efficient graph-based image
  segmentation. International Journal of Computer Vision 59~(2), 167--181.

\bibitem[{Gerke(2015)}]{gerke2015techrepo}
Gerke, M., 2015. Use of the stair vision library within the {ISPRS 2D} semantic
  labeling benchmark {(Vaihingen)}. Tech. rep., ITC, Univ. of Twente.

\bibitem[{Gim\'enez et~al.(2017)Gim\'enez, de~Jong, Peruta, Keller, and
  Schaepman}]{gimenez2017rse}
Gim\'enez, M.~G., de~Jong, R., Peruta, R.~D., Keller, A., Schaepman, M.~E.,
  2017. Determination of grassland use intensity based on multi-temporal remote
  sensing data and ecological indicators. Remote Sensing of Environment 198,
  126 -- 139.

\bibitem[{Golipour et~al.(2016)Golipour, Ghassemian, and
  Mirzapour}]{golipour2016tgrs}
Golipour, M., Ghassemian, H., Mirzapour, F., Feb 2016. Integrating hierarchical
  segmentation maps with mrf prior for classification of hyperspectral images
  in a bayesian framework. IEEE Transactions on Geoscience and Remote Sensing
  54~(2), 805--816.

\bibitem[{Gould et~al.(2008)Gould, Rodgers, Cohen, Elidan, and
  Koller}]{gould2008ijcv}
Gould, S., Rodgers, J., Cohen, D., Elidan, G., Koller, D., 2008. Multi-class
  segmentation with relative location prior. Int. J. Comp. Vision 80~(3),
  300--316.

\bibitem[{Hariharan et~al.(2011)Hariharan, Arbelaez, Bourdev, Maji, and
  Malik}]{hariharan2011iccv}
Hariharan, B., Arbelaez, P., Bourdev, L., Maji, S., Malik, J., 2011. Semantic
  contours from inverse detectors. In: International Conference on Computer
  Vision (ICCV).

\bibitem[{Hariharan et~al.(2015)Hariharan, Arbel\'aez, Girshick, and
  Malik}]{hariharan2015cvpr}
Hariharan, B., Arbel\'aez, P., Girshick, R., Malik, J., 2015. Hypercolumns for
  object segmentation and fine-grained localization. In: {IEEE/CVF}
  International Conference on Computer Vision and Pattern Recognition.

\bibitem[{Hedhli et~al.(2016)Hedhli, Moser, Serpico, and
  Zerubia}]{hedhli2016tgrs}
Hedhli, I., Moser, G., Serpico, S., Zerubia, J., 2016. A new cascade model for
  the hierarchical joint classification of multitemporal and multiresolution
  remote sensing data. IEEE Transactions on Geoscience and Remote Sensing
  54~(11), 6333--6348.

\bibitem[{Hoberg et~al.(2015)Hoberg, Rottensteiner, Queiroz-Feitosa, and
  Heipke}]{hoberg2015tgrs}
Hoberg, T., Rottensteiner, F., Queiroz-Feitosa, R., Heipke, C., 2015.
  Conditional random fields for multitemporal and multiscale classification of
  optical satellite imagery. IEEE Transactions on Geoscience and Remote Sensing
  53~(2), 659--673.

\bibitem[{H\"ohle(2017)}]{hoehle2017remsens}
H\"ohle, J., 2017. Generating topographic map data from classification results.
  Remote Sensing 9~(3).

\bibitem[{Huang et~al.(2017)Huang, Liu, van~der Maaten, and
  Weinberger}]{huang2017cvpr}
Huang, G., Liu, Z., van~der Maaten, L., Weinberger, K.~Q., 2017. Densely
  connected convolutional networks. In: IEEE Conference on Computer Vision and
  Pattern Recognition (CVPR).

\bibitem[{Jat et~al.(2008)Jat, Garg, and Khare}]{jat2008jag}
Jat, M.~K., Garg, P., Khare, D., 2008. Monitoring and modelling of urban sprawl
  using remote sensing and gis techniques. International Journal of Applied
  Earth Observation and Geoinformation 10~(1), 26 -- 43.

\bibitem[{Kluckner et~al.(2009)Kluckner, Mauthner, Roth, and
  Bischof}]{kluckner2009accv}
Kluckner, S., Mauthner, T., Roth, P.~M., Bischof, H., 2009. Semantic
  classification in aerial imagery by integrating appearance and height
  information. In: ACCV 2009, Xi\'an (China).

\bibitem[{Kokkinos(2016)}]{kokkinos2016iclr}
Kokkinos, I., 2016. Pushing the boundaries of boundary detection using deep
  learning. In: International Conference on Learning Representations (ICLR).

\bibitem[{Kokkinos(2017)}]{kokkinos2017cvpr}
Kokkinos, I., 2017. In: {IEEE/CVF} International Conference on Computer Vision
  and Pattern Recognition.

\bibitem[{Krizhevsky et~al.(2012)Krizhevsky, Sutskever, and
  Hinton}]{krizhevsky2012nips}
Krizhevsky, A., Sutskever, I., Hinton, G., 2012. {ImageNet} classification with
  deep convolutional neural networks. In: Advances in Neural Information
  Processing Systems.

\bibitem[{Lafferty et~al.(2001)Lafferty, McCallum, and
  Pereira}]{lafferty2001icml}
Lafferty, J.~D., McCallum, A., Pereira, F. C.~N., 2001. Conditional random
  fields: Probabilistic models for segmenting and labeling sequence data. In:
  Proceedings of the Eighteenth International Conference on Machine Learning.
  ICML '01. pp. 282 -- 289.

\bibitem[{Lagrange et~al.(2015)Lagrange, Le~Saux, Beaupere, Boulch,
  Chan-Hon-Tong, Herbin, Randrianarivo, and Ferecatu}]{lagrange2015igarss}
Lagrange, A., Le~Saux, B., Beaupere, A., Boulch, A., Chan-Hon-Tong, A., Herbin,
  S., Randrianarivo, H., Ferecatu, M., 2015. Benchmarking classification of
  earth-observation data: From learning explicit features to convolutional
  networks. In: IEEE International Geoscience and Remote Sensing Symposium.
  Milan, Italy, pp. 4173 -- 4176.

\bibitem[{LeCun et~al.(1998)LeCun, Bottou, Bengio, and
  Haffner}]{lecun1998pieee}
LeCun, Y., Bottou, L., Bengio, Y., Haffner, P., 1998. Gradient-based learning
  applied to document recognition. Proceedings of the IEEE.

\bibitem[{Lempitsky et~al.(2011)Lempitsky, Vedaldi, and
  Zisserman}]{lempitsky2011nips}
Lempitsky, V., Vedaldi, A., Zisserman, A., 2011. Pylon model for semantic
  segmentation. In: Shawe-Taylor, J., Zemel, R.~S., Bartlett, P.~L., Pereira,
  F., Weinberger, K.~Q. (Eds.), Advances in Neural Information Processing
  Systems 24 (NIPS). pp. 1485 -- 1493.

\bibitem[{Li et~al.(2015)Li, Femiani, Xu, Zhang, and Wonka}]{li2015tgrs}
Li, E., Femiani, J., Xu, S., Zhang, X., Wonka, P., Aug 2015. Robust rooftop
  extraction from visible band images using higher order crf. IEEE Transactions
  on Geoscience and Remote Sensing 53~(8), 4483--4495.

\bibitem[{Long et~al.(2015)Long, Shelhamer, and Darrell}]{long2015cvpr}
Long, J., Shelhamer, E., Darrell, T., 2015. Fully convolutional networks for
  semantic segmentation. In: {IEEE/CVF} International Conference on Computer
  Vision and Pattern Recognition.

\bibitem[{Maggiori et~al.(2017)Maggiori, Tarabalka, Charpiat, and
  Alliez}]{maggiori2017tgrs}
Maggiori, E., Tarabalka, Y., Charpiat, G., Alliez, P., 2017. High-resolution
  aerial image labeling with convolutional neural networks. IEEE Transaction on
  Geoscience and Remote Sensing 55~(12), 7092 -- 7103.

\bibitem[{Maninis et~al.(2017)Maninis, Pont-Tuset, Arbel\'{a}ez, and
  Gool}]{maninis2017tpami}
Maninis, K., Pont-Tuset, J., Arbel\'{a}ez, P., Gool, L.~V., 2017. Convolutional
  oriented boundaries: From image segmentation to high-level tasks. IEEE
  Transactions on Pattern Analysis and Machine Intelligence.

\bibitem[{Marcos et~al.(in press)Marcos, Volpi, Kellenberger, and
  Tuia}]{marcos2018jisprs}
Marcos, D., Volpi, M., Kellenberger, B., Tuia, D., in press. Land cover mapping
  at very high resolution with rotation equivariant {CNN}s: towards small yet
  accurate models. ISPRS Journal of Photogrammetry and Remote Sensing.

\bibitem[{Marmanis et~al.(2018)Marmanis, Schindler, Wegner, Galliani, Datcu,
  and Stilla}]{marmanis2018jisprs}
Marmanis, D., Schindler, K., Wegner, J.~D., Galliani, S., Datcu, M., Stilla,
  U., 2018. Classification with an edge: Improving semantic image segmentation
  with boundary detection. ISPRS Journal of Photogrammetry and Remote Sensing
  135, 158 -- 172.

\bibitem[{Mostajabi et~al.(2015)Mostajabi, Yadollahpour, and
  Shakhnarovich}]{mostajabi2015cvpr}
Mostajabi, M., Yadollahpour, P., Shakhnarovich, G., 2015. Feedforward semantic
  segmentation with zoom-out feature. In: IEEE/CVF International Conference on
  Computer Vision and Pattern Recognition (CVPR).

\bibitem[{Rother et~al.(2007)Rother, Kolmogorov, Lempitsky, and
  Szummer}]{rother2007cvpr}
Rother, C., Kolmogorov, V., Lempitsky, V., Szummer, M., 2007. Optimizing binary
  mrfs via extended roof duality. In: IEEE/CVF Conference on Computer Vision
  and Pattern Recongnition (CVPR).

\bibitem[{Shen et~al.(2015)Shen, Wang, Wang, Bai, and Zhang}]{shen2015cvpr}
Shen, W., Wang, X., Wang, Y., Bai, X., Zhang, Z., 2015. Deepcontour: A deep
  convolutional feature learned by positive-sharing loss for contour detection.
  In: IEEE Conference on Computer Vision and Pattern Recognition (CVPR).

\bibitem[{Sherrah(2016)}]{sherrah2016arxiv}
Sherrah, J., 2016. Fully convolutional networks for dense semantic labelling of
  high-resolution aerial imagery. arXiv:1606.02585.

\bibitem[{Shotton et~al.(2006)Shotton, Winn, Rother, and
  Criminisi}]{shotton2006eccv}
Shotton, J., Winn, J., Rother, C., Criminisi, A., 2006. Textonboost: Joint
  appearance, shape and context modeling for multi-class object recognition and
  segmentation. In: European Conference on Computer Vision.

\bibitem[{Simonyan and Zisserman(2015)}]{simonyan2015iclr}
Simonyan, K., Zisserman, A., 2015. Very deep convolutional networks for
  large-scale image recognition. In: International Conference in Learning
  Representation. Vol. abs/1409.1.

\bibitem[{Uijlings and Ferrari(2015)}]{uijlings2015cvpr}
Uijlings, J. R.~R., Ferrari, V., 2015. Situational object boundary detection.
  In: IEEE Computer Vision and Pattern Recognition (CVPR).

\bibitem[{Volpi and Ferrari(2015)}]{volpi2015bcvprw}
Volpi, M., Ferrari, V., 2015. Semantic segmentation of urban scenes by learning
  local class interactions. In: IEEE/CVF CVPRW Earthvision.

\bibitem[{Volpi and Tuia(2017)}]{volpi2017tgrs}
Volpi, M., Tuia, D., 2017. Dense semantic labeling of subdecimeter resolution
  images with convolutional neural networks. {IEEE Trans. Geosci. Remote Sens.}
  55~(2), 881--893.

\bibitem[{Xie and Tu(2015)}]{xie2015cvpr}
Xie, S., Tu, Z., 2015. Holistically-nested edge detection. In: IEEE/CVF
  International Conference on Computer Vision (CVPR).

\bibitem[{Zheng et~al.(2017)Zheng, Zhang, and Wang}]{zheng2017tgrs}
Zheng, C., Zhang, Y., Wang, L., May 2017. Semantic segmentation of remote
  sensing imagery using an object-based markov random field model with
  auxiliary label fields. IEEE Transactions on Geoscience and Remote Sensing
  55~(5), 3015--3028.

\bibitem[{Zhong and Wang(2007)}]{zhong2007tgrs}
Zhong, P., Wang, R., 2007. A multiple conditional random fields ensemble model
  for urban area detection in remote sensing optical images. IEEE Transaction
  on Geoscience and Remote Sensing 45~(12), 3978--3988.

\end{thebibliography}

\end{document}